\documentclass[fleqn,10pt]{wlscirep}
\usepackage[utf8]{inputenc}
\usepackage[T1]{fontenc}
\usepackage{tcolorbox}
\usepackage{amsmath}
\usepackage{multirow}
\usepackage{pifont}
\usepackage{caption}

\newcommand{\ok}{\ding{51}}      % ✓
\newcommand{\half}{(\ding{51})}  % (✓)
\newcommand{\no}{\ding{55}}      % ✗

\usepackage{tcolorbox} % 导言区引入

\newtcolorbox{mybox}[2][]{colback=gray!10, colframe=gray, 
  fonttitle=\bfseries, coltitle=black,
  title=#2,#1}

\title{Tibetan Language and AI: A Comprehensive Survey of Resources, Methods and Challenges}

\author[1,3,+]{Cheng Huang}
\author[2,+,*]{Nyima Tashi}
\author[1]{Fan Gao}
\author[1]{Yutong Liu}
\author[4]{Jiahao Li}
\author[5]{Hao Tian}
\author[6]{Siyang Jiang}
\author[1,2]{Thupten Tsering}
\author[2]{Ban Ma-bao}
\author[2]{Renzeg Duojie}
\author[2]{Gadeng Luosang}
\author[2]{Rinchen Dongrub}
\author[2]{Dorje Tashi}
\author[1]{Jin Zhang}
\author[1]{Xiao Feng}
\author[1,7]{Hao Wang}
\author[8]{Jie Tang}
\author[9,10]{Guojie Tang}
\author[1,*]{Xiangxiang Wang}
\author[3]{Jia Zhang}
\author[12]{Tsengdar Lee}
\author[1,*]{Yongbin Yu}

\affil[1]{University of Electronic Science and Technology of China, Chengdu, 610056, China}
\affil[2]{Tibet University, Lhasa 850000, China}
\affil[3]{Southern Methodist University, Dallas, 75205, USA}
\affil[4]{The City University of Hong Kong, Kowloon, Hong Kong SAR 999077, China}
\affil[5]{The Hong Kong Polytechnic University, Hung Hom, Kowloon, Hong Kong SAR 999077, China}
\affil[6]{The Chinese University of Hong Kong, Sha Tin, Hong Kong SAR 999077, China}
\affil[7]{University of Connecticut, Storrs, 06269, USA}
\affil[8]{Tsinghua University, Beijing 100084, China}
\affil[9]{University of Texas at Arlington, Arlington, TX 76019, USA}
\affil[10]{University of Chinese Academy of Sciences, Beijing 100049, China}
\affil[11]{The State Key Laboratory of Tibetan Intelligence, Lhasa 850000, China}
\affil[12]{National Aeronautics and Space Administration, Washington, DC 20546, USA}
\affil[*]{Corresponding Author, E-mail: ybyu@uestc.edu.cn}
\affil[+]{Equal Contribution}

\begin{abstract}
Tibetan, one of the major low-resource languages in Asia, presents unique linguistic and sociocultural characteristics that pose both challenges and opportunities for AI research. Despite increasing interest in developing AI systems for underrepresented languages, Tibetan has received limited attention due to a lack of accessible data resources, standardized benchmarks, and dedicated tools. This paper provides a comprehensive survey of the current state of Tibetan AI in the AI domain, covering textual and speech data resources, NLP tasks, machine translation, speech recognition, and recent developments in LLMs. We systematically categorize existing datasets and tools, evaluate methods used across different tasks, and compare performance where possible. We also identify persistent bottlenecks such as data sparsity, orthographic variation, and the lack of unified evaluation metrics. Additionally, we discuss the potential of cross-lingual transfer, multi-modal learning, and community-driven resource creation. This survey aims to serve as a foundational reference for future work on Tibetan AI research and encourages collaborative efforts to build an inclusive and sustainable AI ecosystem for low-resource languages.
\end{abstract}
\begin{document}

\flushbottom
\maketitle
%\thispagestyle{empty}

%\noindent Please note: Abbreviations should be introduced at the first mention in the main text - no abbreviations lists. Suggested structure of main text (not enforced) is provided below.

\begin{mybox}{BOX 1: Key Contributions of This Survey}
\label{box-1}
\begin{itemize}
    \item Presents the first unified survey of Tibetan AI across text, speech, and emerging multimodal domains, rather than treating them as isolated research tracks.
    \item Systematically consolidates all publicly accessible Tibetan resources: datasets, toolchains, benchmarks, and linguistic assets to provide a structured foundation for future development.
    \item Critically analyzes the behavior of state-of-the-art AI techniques under Tibetan-specific linguistic and sociocultural conditions, revealing failure modes not observed in high-resource settings.
    \item Distills the remaining research gaps and proposes a forward-looking agenda toward inclusive, sustainable, and culturally grounded Tibetan AI research.
\end{itemize}

\end{mybox}

\noindent \textbf{Note}: Some abbreviations are summaried in Appendix~\ref{app1}~Table~\ref{app-1}. These abbreviations represent some of the classical or key concepts, methods, and datasets frequently mentioned throughout this survey.

\clearpage

\begin{figure*}[ht]
    \centering
    \includegraphics[width=\linewidth]{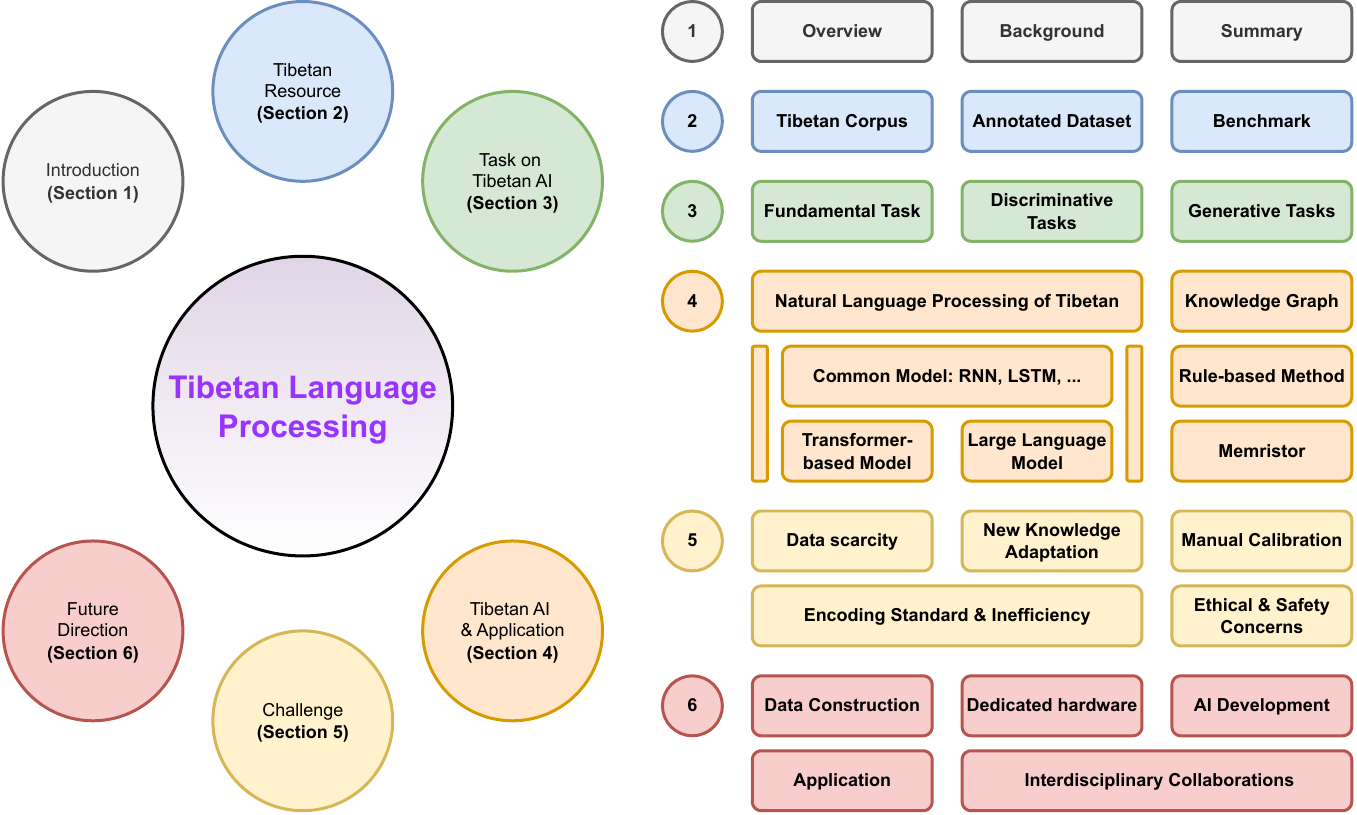}
    \caption{Overview of the Practical Guide for current AI Trend on Tibetan}
    \label{in-1}
\end{figure*}

\section{Introduction}
\label{s1}

In recent years, AI has achieved remarkable progress in language understanding\cite{bert,attention,roberta,albert,xlnet}, generation\cite{bart,pegasus,seq2seq,neural,gpt2}, and multimodal interaction\cite{vilbert,visualbert,clip,blip2,flamingo}, largely driven by advances in deep learning\cite{bert,tibert}, the availability of large-scale datasets\cite{tibstccot,TIB-STC,TJUNLP}, and the emergence of LLMs, such as GPT family\cite{gpt4,gpt4o,gpt5}, LlaMA family\cite{llama,llama2,llama3}, DeepSeek family\cite{r1,v3}, Qwen family\cite{qwen,qwenv2,qwenv2.5,qwenv3}, Gemini family\cite{gemini1,gemini1.5,gemini2}, and Claude family\cite{claude,claude3,claude35,claude4}. While such progress has significantly benefited high-resource languages such as English and Chinese, many languages spoken by marginalized and linguistically diverse communities, commonly referred to as low-resource languages, remain underrepresented in the development and deployment of language technologies. Among them, Tibetan is a particularly important yet underexplored case. It is a member of the Tibeto-Burman branch of the Sino-Tibetan language family, is spoken by over six million people across China, India, Nepal, and Bhutan\cite{bg-1}. The language plays a crucial role in preserving the cultural, spiritual, and historical heritage of the Himalayan region\cite{bg-2,bg-3}. However, Tibetan exhibits several characteristics that complicate computational modeling: a non-Latin orthography, morpho-syllabic structure, complex verb morphology, dialectal variation, and limited digital standardization\cite{bg-4,bg-5}. These features, combined with a lack of accessible linguistic resources and annotated data, make Tibetan a challenging target for current AI systems\cite{tlue,TIB-STC,tibstccot}.

Despite the increasing academic interest in developing language technologies for low-resource and endangered languages, Tibetan remains relatively neglected in both academic and industrial research. Existing efforts are fragmented across isolated datasets, task-specific methods, and small-scale experiments. There is no unified survey or roadmap that summarizes the current landscape of Tibetan AI research, identifies key gaps, and proposes a direction for future development. This absence hinders collaboration, benchmarking, and progress. These research gaps motivate this review which offers a comprehensive review of the development of AI technologies and their applications in Tibetan. We aim to cover topics on existing AI models, various tasks of AI on Tibetan, AI applications of Tibetan language, arising challenges, and future directions.

As shown in Figure~\ref{in-1}, this review seeks to answer the following questions:
\begin{itemize}
    \item \textbf{Section~\ref{s1}:} What is the current status of cross-disciplinary research on AI and Tibetan? What have we summarized in this report?
    \item \textbf{Section~\ref{s2}:} What is Tibetan? What are the unique characteristics of Tibetan compared to major languages such as Chinese and English? How can we obtain relevant data?
    \item \textbf{Section~\ref{s3}:} What are the main tasks for existing AI applications in Tibetan? What are the needs for these tasks?
    \item \textbf{Section~\ref{s4}:} What AI-related technologies do they use and how is the progress? How are they set up according to the needs of the Tibetan language itself?
    \item \textbf{Section~\ref{s5}:} What challenges should be addressed when implementing AI in Tibetan?
    \item \textbf{Section~\ref{s6}:} How can we optimize the construction
 of AI to enhance their applicability in Tibetan AI, ultimately contributing to low-resource language and creating a positive
 societal impact?
\end{itemize}

To address these issues, this survey provides the first comprehensive review of Tibetan language research in the field of AI. We begin by outlining the current research landscape at the intersection of Tibetan and AI, highlighting the key challenges that hinder progress in this low-resource domain. Building on this context, we articulate the motivation for conducting this survey and provide an overview of its themes and structure in Section~\ref{s1}. In Section~\ref{s2}, we systematically review available data resources for both text and speech, covering monolingual corpora, parallel corpora, and annotated datasets for tasks such as word segmentation, part-of-speech tagging, and named entity recognition. Section~\ref{s3} summarizes key tasks at the intersection of AI and Tibetan, analyzing their practical needs and applications. In Section~\ref{s4}, we examine AI models applied to Tibetan, including machine translation, classification, automatic speech recognition, and text-to-speech synthesis. We further assess the performance and limitations of multilingual pre-trained models and LLMs when applied to Tibetan, and explore emerging directions such as cross-lingual transfer, prompt-based learning, and multimodal processing. Section~\ref{s5} then analyzes the bottlenecks of existing AI models and data, and discusses ethical and safety considerations. Finally, Section~\ref{s6} outlines promising avenues for future research, including the development of specialized hardware for Tibetan scripts, the construction of larger and more diverse datasets and benchmarks, the advancement of multimodal models, and the exploration of techniques such as instruction tuning and reasoning-augmented learning. These efforts can enhance the efficiency of Tibetan AI, broaden task coverage, support applications in education and cultural preservation, and help bridge the gap between low- and high-resource languages, thereby promoting inclusivity in multilingual AI.

We hope this work will serve as a foundational reference for researchers and practitioners in low-resource AI, and contribute to the broader goal of equitable language technology development.

\clearpage

\begin{mybox}{BOX 2: Background of Tibetan}
\label{box-2}

Tibetan is a historically rich and linguistically diverse language with a distinctive script, a strong classical literary tradition, and significant dialectal variation, playing a central role in the cultural and religious life of communities across the Himalayan region\cite{bg-3,bg-1}.

\textbf{$\to$ Historical Origin} \\Tibetan belongs to the Tibeto-Burman branch of the Sino-Tibetan language family. The development of written Tibetan began in the 7th century during the reign of King Songtsen Gampo, who commissioned the creation of the Tibetan script. This script, attributed to Thonmi Sambhota and based on Indian Brahmi and Gupta models, was primarily designed to support the translation of Buddhist texts from Sanskrit, laying the foundation for a written literary tradition\cite{bg-6,bg-8}.

\textbf{$\to$ Script \& Classical Literary Tradition} \\The Tibetan script is an abugida system composed of consonant characters combined with vowel diacritics. As shown in Figure~\ref{box-1-f-1}, this is one sample of the native tibetan syllable structure, which means raise or lift. It consists of 7 parts: prefix letter, root letter, superscript letter, subscript letter, vowel letter, suffix letter, and farther suffix letter. There is a certain combination principle between different letters\cite{bg-5}.
Since its inception, it has remained relatively stable and is still used in both religious and secular contexts. Classical Tibetan emerged as the standard literary form and became the medium for a vast body of canonical Buddhist literature, historical records, legal documents, and medical texts. This classical form continues to serve as a common written standard across Tibetan-speaking regions despite spoken variation\cite{bg-9}.

\begin{center}
  \includegraphics[width=0.75\linewidth]{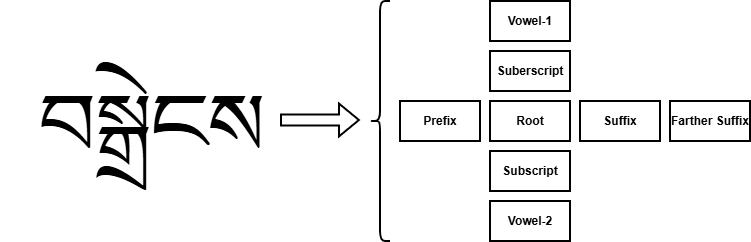} 
  \captionof{figure}{One Structure Example of the Tibetan Script}
  \label{box-1-f-1}
\end{center}

\textbf{$\to$ Dialectal Diversity} \\Modern spoken Tibetan is generally divided into three major dialect groups: Central (Ü-Tsang), Khams, and Amdo. These dialects exhibit significant differences in phonology, vocabulary, and syntax, and are often mutually unintelligible. However, they are unified by a shared use of the classical written language. Among them, Central Tibetan, especially the Lhasa variety, serves as the basis for standardized written Tibetan and official media usage\cite{bg-10,bg-9}.

\textbf{$\to$ Contemporary Use \& Sociolinguistic Role} \\Today, Tibetan is spoken by more than six million people across the Tibet Autonomous Region of China, as well as in parts of India, Nepal, and Bhutan. It is actively used in domains such as religious practice, local education, community governance, broadcasting, and literature. Tibetan remains a vital symbol of cultural identity, continuity, and resistance, especially among communities in exile and in regions experiencing language shift or political pressur\cite{bg-11,bg-12}.

\end{mybox}

\section{Tibetan Resource}
\label{s2}

Despite its rich cultural and linguistic heritage, shown in Box~2, Tibetan has long suffered from a lack of standardized and accessible language resources. In this section, we provide an overview of existing Tibetan resources, including public corpora, annotated datasets, and preliminary benchmarks, which together form the foundation for Tibetan AI and its research.

\subsection{Public Corpus}
Existing Tibetan public corpora primarily consist of web-crawled texts, classical scriptures, news articles, and limited modern-domain content. While these corpora provide a crucial starting point for data-driven Tibetan language processing, most of them suffer from uneven quality, inconsistent orthographic standards, duplicated content, and a lack of linguistic normalization. Furthermore, a large portion of publicly available Tibetan text is derived from religious or historical sources, which introduces strong stylistic bias and limits the applicability of these corpora for training models intended for contemporary usage scenarios such as dialogue, instruction following, and safety alignment. To better understand the foundation upon which later datasets are constructed, it is necessary to first examine the open Tibetan language databases that supply the raw materials for corpus creation. Unlike task-specific datasets or benchmarks, a number of open Tibetan language repositories exist in the form of large-scale raw text collections and lexical platforms. These databases provide the foundational materials from which corpora and downstream AI datasets are later derived.

\begin{itemize}
    \item \textbf{BDRC}\cite{BDRC}: BDRC is a large-scale digital archive containing over 200k volumes of Tibetan canonical texts in PDF and Unicode formats. It is the largest publicly accessible Tibetan text source but heavily biased toward classical and religious style.
    \item \textbf{THL}\cite{THL}: THL is a curated linguistic platform providing lexical standards, dialect documentation, orthographic normalization guides, and annotated language samples. It is widely used as a reference for tokenization and script normalization rather than direct model training.
    \item \textbf{RYTC}\cite{rytc}: RYTC is a structured Tibetan Buddhist corpus with semantic segmentation and commentary alignment. While domain-constrained, its clean formatting makes it suitable for canonical language modeling.
    \item \textbf{BDICT/THL Dictionary API}\cite{THLDictionary}: THL Dictionary API is an openly accessible lexical database offering lemma forms, POS tags, and example sentences with API support. It serves as a grounding resource for morphological analysis and vocabulary construction in Tibetan AI.
    \item \textbf{TWD}\cite{TibetanWiki}: TWD is a modern-domain encyclopedic corpus with around 70k+ Tibetan articles. Although limited in size, it remains one of the very few sources that reflect contemporary Tibetan usage suitable for instruction tuning and conversational modeling.
    \item \textbf{ACIP}\cite{ACIP}: ACIP is An early digitization initiative encoding Tibetan texts using Wylie transliteration and Unicode mapping. It plays a foundational role in script standardization and character-level normalization.
    \item \textbf{OTFC}\cite{OpenSubtitlesTibetan}: OTFC is a small but unique set of Tibetan subtitle fragments that contain informal and conversational speech patterns, valuable for modeling colloquial Tibetan despite its extremely limited scale.
\end{itemize}

These repositories form the base layer of the Tibetan language resource stack and serve as the raw material for constructing standardized corpora, task datasets, and eventually benchmarks for Tibetan AI.

\subsection{Annotated Dataset}
Compared to raw corpora and lexical repositories, publicly available Tibetan annotated datasets are extremely limited in both scale and task diversity. Most existing resources were constructed for early AI tasks such as word segmentation, POS tagging, and OCR correction, often derived from classical texts with manually curated labels. These datasets, while valuable for linguistic analysis, lack consistency in annotation standards and are not aligned with modern instruction-based or safety-critical AI development needs. Furthermore, there is currently no unified annotation framework that supports multi-dimensional supervision such as instruction-response pairs, safety alignment labels, or preference signals, which are essential for training contemporary Tibetan LLMs. To better understand the current landscape of Tibetan annotated resources and their limitations in relation to LLM-oriented supervision, we summarize the main publicly available datasets as follows:
\begin{itemize}
    \item \textbf{NICT-Tib1}\cite{NICTTib1}: NICT-Tib1 comprises approximately 33.5 hours of read speech from native Lhasa Tibetan speakers and has been widely adopted for ASR model training, although its scripted design limits variability in prosody and spontaneous speech patterns.
    
    \item \textbf{TIBMD@MUC}\cite{TIBMDMUC}: TIBMD@MUC covers three major dialects: Ü-Tsang, Kham, and Amdo, to introduce acoustic diversity into Tibetan ASR research, yet it remains limited to speech transcription without incorporating semantic or generative supervision.
    
    \item \textbf{TibNER}\cite{TibNER}: TibNER provides token-level entity annotations over person, location, and organization categories and supports named entity recognition research, but it does not extend to instruction-based reasoning or generative entity understanding.
    
    \item \textbf{CUTE Multilingual Corpus}\cite{CUTECorpus}: CUTECorpus introduces Tibetan into a multilingual alignment setting alongside Chinese, Uyghur, and English and enables parallel translation research, though it lacks instruction tuning or preference alignment signals.
    
    \item \textbf{Tibetan MRC}\cite{TibetanMRC}: TibetanMRC pairs Tibetan news passages with manually created question, answer annotations and advances extractive reading comprehension but does not incorporate dialogue context or chain-of-thought reasoning supervision.
    
    \item \textbf{TiConvQA}\cite{TiConvQA}: TiConvQA extends Tibetan QA into multi-turn conversational form and enriches discourse-level information retrieval, yet it still follows a fixed QA paradigm without integrating preference or safety alignment annotations.
    
    \item \textbf{TiLTS}\cite{TiLTS}: TiLTS aligns Tibetan news articles with abstractive summaries and introduces long-form generative modeling, but it remains decoupled from controllable instruction-dependent generation objectives.
    
    \item \textbf{Cross-Lingual Tibetan Summarization}\cite{CrossLingualSummarization}: CrossLingualSummarization pairs Tibetan source passages with human-written Chinese summaries and supports bilingual summarization research, although it focuses exclusively on content fidelity without multi-dimensional supervision signals.
    
    \item \textbf{TiKG-30K}\cite{TiKG30K}: TiKG-30K provides over 30K Tibetan entity-relation triples and facilitates symbolic reasoning, yet it does not connect structured knowledge supervision with instruction-based or conversational LLM training.
\end{itemize}

Also, for contemporary LLM development, publicly available Tibetan pretraining datasets are extremely scarce, with only three large-scale corpora currently accessible to the community:
\begin{itemize}
    \item \textbf{TIBSTC}\cite{TIB-STC}: TIB-STC introduces the first large-scale, expert-curated and structurally organized Tibetan corpus designed specifically for LLM development, consolidating over 11B tokens across diverse domains such as literature, religion, medicine, law, and daily communication. Unlike earlier fragmented Tibetan resources, TIB-STC establishes a unified supervision framework that supports pretraining, instruction tuning, and alignment phases, and is further validated through a three-stage pipeline language modeling, supervised fine-tuning, and preference optimization, using the reference model Sun-Shine. To enable systematic evaluation, the authors also align the corpus with TLUE\cite{tlue}, a Tibetan benchmark suite covering understanding, reasoning, and safety scenarios, and demonstrate significant performance improvements on tasks such as Ti-MMLU and Ti-SafetyBench. TIB-STC therefore marks the first comprehensive and publicly reported attempt to construct a scalable Tibetan data infrastructure for modern LLM training and evaluation.
    \item \textbf{TIBSTC-CoT}\cite{tibstccot}: TIBSTC-CoT introduces the first chain-of-thought oriented Tibetan instruction corpus, extending the TIBSTC framework by generating multi-domain reasoning data through CoT prompting to address the scarcity of logical supervision in low-resource Tibetan language modeling. The dataset covers diverse reasoning scenarios, including arithmetic reasoning, commonsense inference, and structured logical deduction and follows a scalable, replicable data generation pipeline that leverages large multilingual models to produce Tibetan step-by-step reasoning trajectories. Built on this corpus, the authors train the Sunshine-thinking model family, which demonstrates strong performance on Tibetan reasoning tasks and reaches competitive results compared to state-of-the-art multilingual LLMs, highlighting the effectiveness of explicit CoT supervision. TIBSTC-CoT therefore establishes the first dedicated Tibetan reasoning dataset and provides a practical methodology for injecting structured cognitive signals into low-resource language models.
    \item \textbf{TJUNLP Tibetan Corpus}\cite{TJUNLP}: TJUNLP Tibetan Corpus establishes the largest curated Tibetan pretraining resource to date, consolidating 72GB of Tibetan text from open-source corpora, web-crawled pages, synthetic translation pairs, and private digitized collections under a unified cleaning and deduplication pipeline tailored to Tibetan linguistic characteristics. Built to support continual pretraining of multilingual foundation models, the corpus enables two-stage adaptation, language-balanced pretraining and long-context extension, achieving substantial performance gains across Tibetan benchmarks when integrated into models such as Qwen2.5\cite{qwenv2.5}. The work demonstrates that large-scale, well-filtered Tibetan pretraining data can significantly enhance both understanding and generation capabilities and serves as the current backbone corpus for Tibetan LLM training under the TJUNLP Tibetan Corpus designation.
\end{itemize}

Together, these three corpora delineate the current landscape of Tibetan LLM data resources: TIBSTC\cite{TIB-STC} establishes the foundation for supervised instruction alignment, TIBSTC-CoT\cite{tibstccot} contributes explicit reasoning supervision, and the TJUNLP Tibetan Corpus\cite{TJUNLP} supplies the large-scale pretraining backbone necessary for continual Tibetan language modeling. For more details about TIBSTC\cite{TIB-STC} and TIBSTC-CoT\cite{tibstccot}, please refer to Appendix \ref{app2}.

\subsection{Benchmark}

Both dedicated, formal and official Tibetan benchmarks and multilingual evaluation suites with meaningful Tibetan coverage are exceedingly limited, highlighting a significant evaluation gap in Tibetan language AI. To provide a structured analysis, we divide existing evaluation resources into two categories: \textit{Specialized Benchmark} and \textit{Comprehensive Multilingual Benchmark} with partial Tibetan inclusion. We introduce them separately in the following.

\subsubsection{Specialized Benchmark}

Despite the rapid development of multilingual evaluation resources, Tibetan remains one of the most underserved languages, with only a few benchmarks that are still actively adopted by the research community. The currently recognized evaluation resources can be categorized into two representative benchmarks:

\begin{itemize}
\item \textbf{TLUE}\cite{tlue}: TLUE is the first and currently the only standardized benchmark designed specifically for evaluating natural language understanding in Tibetan for LLMs. TLUE integrates Ti-MMLU and Ti-SafetyBench to jointly assess general knowledge reasoning and safety alignment, with support for zero-shot and few-shot settings to enable fair comparisons without language-specific tuning. More details are provided in Appendix~\ref{app3}.

\item \textbf{Tibetan ASR}\cite{asr}: Tibetan ASR was built primarily upon the Tibetan portion of the IARPA Babel Corpus, which contains around 40 hours of transcribed conversational speech under low-resource conditions. Although originally part of a multilingual ASR initiative, it has become a de facto standalone benchmark for Tibetan speech recognition, with Word Error Rate as the standard metric, representing one of the earliest task-specific benchmarks for Tibetan spoken language processing.
\end{itemize}

\subsubsection{Comprehensive Multilingual Benchmark}

Among all existing multilingual benchmarks, only FLORES-200\cite{flores} provides limited Tibetan coverage and exclusively in translation, while no general-purpose benchmark assesses Tibetan reasoning, understanding, or safety alignment. To illustrate the limitations of current multilingual resources, we summarize the Tibetan component of FLORES-200\cite{flores} as follows:
\begin{itemize}
    \item \textbf{Tibetan Sub-benchmark of FLORES-200}\cite{flores}: this is a multilingual machine translation benchmark released by Meta AI, covering 200 languages under a unified evaluation protocol, and Tibetan (bo) is included as one of the low-resource tracks for Tibetan-English and Tibetan-Chinese translation using BLEU\cite{bleu} and ChrF++\cite{chrf} as metrics. However, the Tibetan portion of FLORES-200 is strictly limited to translation and does not include any tasks related to reasoning, dialogue, safety alignment, or domain-specific Tibetan linguistic understanding. The evaluation sentences are generic and do not reflect challenges unique to Tibetan, such as its agglutinative morphology, religious terminology, and orthographic variability.  
\end{itemize}

Although FLORES-200\cite{flores} is the only multilingual benchmark that officially includes Tibetan, it cannot be considered a dedicated Tibetan understanding benchmark, as multilingual models that achieve strong performance on high-resource FLORES tracks exhibit a sharp performance drop on the Tibetan subset. This reflects the token-level sparsity and structural underrepresentation of Tibetan in existing multilingual modeling pipelines, reinforcing the necessity of specialized evaluation resources such as TLUE\cite{tlue}. Likewise, while models like mT5\cite{mt5}, BLOOM\cite{bloom}, XGLM\cite{xglm}, and LLaMA-2\cite{llama2} include Tibetan tokens from sources such as Wikipedia or religious web text, this only implies superficial token exposure rather than meaningful evaluation support, since none of these models are trained or benchmarked under a Tibetan-specific task protocol or standardized test set.

\subsection{Others}
In addition to officially released datasets and standardized benchmarks, there also exists a wide range of non-official, community-curated Tibetan resources that are not formally recognized as benchmarks or datasets but are actively used in practical research settings. For more details, please refer to Hugging Face\footnote{\url{https://huggingface.co/datasets?sort=trending&search=tibetan}} and Kaggle\footnote{\url{https://www.kaggle.com/}}.

\clearpage

\begin{figure}[!ht]
    \centering
    \includegraphics[width=0.75\linewidth]{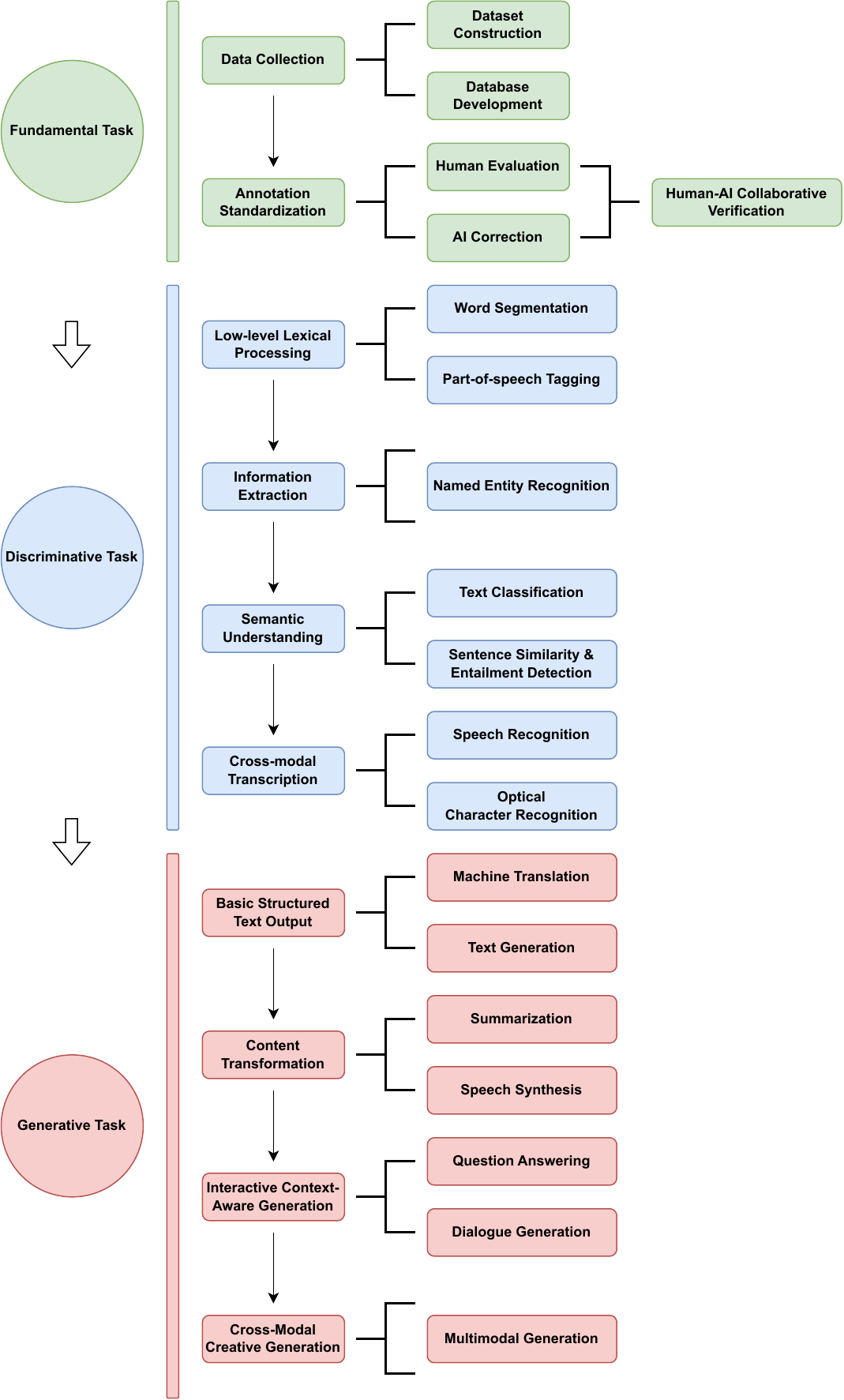}
    \caption{Task Taxonomy of Tibetan AI}
    \label{task-1}
\end{figure}

\section{Task Taxonomy of Tibetan AI}
\label{s3}

This section introduces the task taxonomy of Tibetan AI, detailing each category and explaining their interdependencies within the broader research landscape. Tibetan AI tasks can be organized into three major categories: \textit{Fundamental Task}, \textit{Discriminative Task} and \textit{Generative Task}. As shown in Figure~\ref{task-1}, these three categories themselves form a progressive pipeline, where fundamental tasks provide the necessary resources for discriminative modeling, which in turn underpins more advanced generative applications. We will introduce details about them in following part.

\subsection{Fundamental Task}

Fundamental tasks in Tibetan AI establish the data and resource infrastructure necessary for downstream modeling. For a low-resource and morphologically complex language like Tibetan, this includes collecting and cleaning text and speech corpora, developing balanced domain-specific datasets, standardizing annotation schemes, and building searchable linguistic databases. These efforts, such as digitizing manuscripts, aligning parallel texts, and enabling encoding conversion—are not merely preliminary steps but continuous processes that determine the scalability and effectiveness of all subsequent Tibetan AI research.

\subsubsection{Data Collection}
Data collection serves as the foundational stage for Tibetan AI development, establishing the raw material upon which all downstream modeling relies. It encompasses not only the large-scale acquisition of Tibetan text and speech data, but also the construction of structured databases that enable efficient access, retrieval, and standardized integration of heterogeneous resources. To fulfill these objectives, current efforts in Tibetan AI can be broadly divided into two complementary directions: dataset construction, which focuses on acquiring and cleaning raw multimodal data, and database development, which emphasizes structured organization and standardized storage for long-term reuse.

\begin{itemize}
    \item \textbf{Dataset Construction}: dataset construction is the cornerstone of Tibetan AI, involving the systematic collection, digitization, and preprocessing of Tibetan text and speech data. This includes sourcing data from classical literature, religious scriptures, news articles, educational materials, and modern conversational sources. For speech, high-quality recordings with phoneme-level transcriptions are crucial. Data cleaning steps, such as removing OCR errors, normalizing encoding, and filtering noisy samples, are essential to ensure usability for modeling\cite{liu2025listening}.
    \item \textbf{Database Development}: database development focuses on organizing Tibetan linguistic resources into structured, queryable formats. This includes building searchable lexical databases, parallel corpora repositories, and domain-specific term banks. Databases should support standardized metadata, facilitate interoperability across research projects, and provide APIs or tools for programmatic access\cite{tispell,fmsd}.
\end{itemize}

\subsubsection{Annotation Standardization}

Annotation standardization is essential for ensuring cross-project consistency and enabling interoperability between Tibetan AI systems. Given the linguistic complexity of Tibetan, characterized by syllable-based morphology, script variations, and domain-dependent terminology, a unified annotation protocol is required to harmonize orthographic normalization, part-of-speech tagging, named entity categorization, and segmentation rules. To achieve both scalability and linguistic fidelity, current annotation workflows are increasingly adopting a multi-stage verification paradigm that integrates human expertise with AI-assisted refinement through three complementary processes: 
\begin{itemize}
    \item \textbf{Human Evaluation}: expert annotators manually review and validate linguistic labels to ensure cultural and grammatical fidelity, especially in domains such as religious texts and traditional Tibetan legal terminology\cite{tibstccot,TIB-STC,gpt4o,llama3}.
    \item \textbf{AI Correction}: automated consistency checking is performed using model-assisted detection of label conflicts, segmentation drift, and orthographic variance to reduce manual workload and improve annotation throughput\cite{tibstccot}.
    \item \textbf{Human-AI Collaborative Verification}: a hybrid verification workflow is adopted in which AI-proposed corrections are selectively confirmed or rejected by human experts, allowing scalable annotation while preserving high linguistic precision\cite{tibstccot}.
\end{itemize}

\subsection{Discriminative Task}

Discriminative tasks are foundational to Tibetan AI, particularly in the early stages of system development. These tasks are typically framed as classification or sequence labeling problems, where the model is trained to distinguish between different categories or assign labels to input units. Despite the apparent simplicity, these tasks are particularly challenging in Tibetan due to its agglutinative morphology, lack of whitespace, and dialectal diversity.

\subsubsection{Low-level Lexical Processing}
Low-level lexical processing establishes the token-level foundation for Tibetan language modeling. Due to its syllable-based script and agglutinative morphology, Tibetan lacks clear word boundaries, making segmentation and POS tagging not just preprocessing steps but core computational challenges that directly impact all downstream tasks.

\begin{itemize}
\item \textbf{Word Segmentation}: due to the absence of explicit word boundaries in Tibetan, where text is written as a continuous syllable stream separated only by the tsheg mark ($\cdot$), word segmentation becomes a core prerequisite for all downstream tasks\cite{PW-13,PW-14,ocr-2}. Ambiguous word boundaries, rich morphology, and dialectal variation make segmentation a linguistically complex and computationally challenging problem, directly influencing tasks such as POS tagging, NER, parsing, and machine translation\cite{ProSubCINO}.
\item \textbf{Part-of-speech Tagging}: POS tagging assigns grammatical categories to tokens and serves as a structural layer for higher-level tasks including parsing and translation\cite{fmsd,thl_postagger}. Tibetan’s agglutinative morphology and lack of clear word delimiters make POS tagging heavily dependent on accurate segmentation. The limited availability of standardized tagsets and annotated corpora further complicates model design and evaluation, keeping POS tagging as a critical but under-resourced component in Tibetan AI\cite{tibetan-postcorr}.
\end{itemize}

\subsubsection{Information Extraction}

To extend beyond segmentation and POS tagging, lexical processing in Tibetan also requires recognizing semantically meaningful units such as named entities:

\begin{itemize}
\item \textbf{Named Entity Recognition}: NER involves identifying and classifying entities in text, such as person names, locations, organizations, and temporal expressions, into predefined categories\cite{Tibetan-BERT-wwm}. In Tibetan, NER is essential for information extraction, knowledge graph construction, and question answering. However, the task is complicated by the absence of word boundaries, frequent use of honorifics and case markers, and inconsistency in orthography across domains\cite{TibNER}. The lack of large-scale annotated corpora and the variability of entity naming conventions in religious, historical, and modern contexts further increase difficulty, making Tibetan NER a persistent low-resource challenge\cite{tibetan-bert-ner}.
\end{itemize}

\subsubsection{Semantic Understanding}
Beyond lexical and syntactic processing, semantic understanding focuses on interpreting meaning and relational intent in Tibetan text, enabling higher-level reasoning and downstream applications such as retrieval, dialogue, and alignment. Core tasks in this category include:

\begin{itemize}
\item \textbf{Text Classification}: Tibetan text classification supports applications such as document categorization, sentiment analysis, and topic detection\cite{PW-12,tibert}. While model-agnostic in nature, performance is highly dependent on accurate segmentation and the availability of labeled data. Key challenges include vocabulary sparsity, domain-specific terminology in religious and medical texts, and the lack of consistently annotated public datasets. Recent work explores deep neural classifiers and multilingual transformer models, showing promising results under low-resource conditions.
\item \textbf{Sentence Similarity \& Entailment Detection}: these semantic tasks evaluate relational meaning between Tibetan sentence pairs, supporting QA, retrieval, and paraphrase detection\cite{tibetanruleseg}. Research is limited by the absence of Tibetan contrastive sentence pair datasets and semantic variation arising from honorific expressions, flexible word order, and rich morphology. Existing multilingual models provide partial transfer but lack robust Tibetan grounding, underscoring the need for tailored datasets and alignment strategies for semantic understanding\cite{cino-1,cino}.
\end{itemize}

\subsubsection{Cross-modal Transcription}
While most Tibetan AI research focuses on text-only processing, cross-modal transcription aims to bridge speech, image, and text modalities, enabling large-scale digitization and multimodal understanding of Tibetan linguistic resources. This direction primarily includes two key tasks:

\begin{itemize}
\item \textbf{Speech Recognition}: Tibetan ASR has practical value for voice interfaces, educational tools, oral tradition preservation, and accessibility\cite{fmsd,tmd}. However, development remains limited due to the lack of transcribed speech corpora, strong dialectal variation, and the absence of standardized phoneme inventories. Tibetan’s syllable-based script and tonal patterns further complicate acoustic modeling. While end-to-end architectures such as CTC-based and attention-driven encoder-decoder models have shown initial progress, performance remains unsatisfactory, and the creation of large, domain-diverse Tibetan speech datasets is a major bottleneck.
\item \textbf{Optical Character Recognition}: OCR is essential for digitizing Tibetan manuscripts, scriptures, and printed materials\cite{bg-6}. Progress is challenged by complex orthography with stacked consonants, multiple diacritics, and inconsistent font conventions. Historical documents introduce additional noise such as ink bleed and non-standard layouts. Modern Tibetan OCR increasingly adopts CNN\cite{cnn} and CRNN-style neural architectures\cite{crnn}, but system quality remains constrained by the scarcity of annotated image-text pairs and the absence of standardized benchmarks.
\end{itemize}

\subsection{Generative Task}

Generative tasks in Tibetan AI focus on producing novel outputs, such as text, speech, or structured sequences, based on given input. Unlike discriminative tasks that only assign labels, generative tasks require models to understand context and synthesize coherent, fluent content, making them inherently more complex and resource-intensive. Progress in this area is constrained by limited training data and the absence of standardized evaluation protocols. However, recent advances in multilingual LLMs and instruction-tuned architectures have enabled promising developments across machine translation, summarization, question answering, and speech synthesis for Tibetan.

\subsubsection{Basic Structured Text Output}

At the foundational level of generative modeling, basic structured text output focuses on producing well-formed Tibetan sentences conditioned on external input, ranging from cross-lingual translation to monolingual generation. Current research in this direction primarily centers on two representative tasks:

\begin{itemize}
\item \textbf{Machine Translation}: MT is one of the most studied generative tasks in Tibetan AI, primarily targeting Tibetan-Chinese and Tibetan-English translation\cite{mt-1,mt-2}. Progress is constrained by the scarcity of large-scale parallel corpora, orthographic inconsistency, and domain-specific terminology in religious, legal, and historical texts. While early work relied on rule-based and SMT approaches, recent research has shifted to Transformer-based and mBART-style architectures\cite{attention,bert}, with multilingual pretraining and transfer learning showing promising gains. Initiatives such as NLLB\cite{nllb} and TIB-STC\cite{TIB-STC} attempt to address data bottlenecks, yet Tibetan MT still trails behind other low-resource languages in fluency, robustness, and domain generalization.
\item \textbf{Text Generation}: Tibetan text generation spans story writing, headline generation, dialogue completion, and instruction-response synthesis\cite{tibstccot,TIB-STC}. Challenges arise from limited monolingual corpora, morphological complexity, and strong stylistic variation across literary, religious, and colloquial registers. Multilingual language models such as mT5\cite{mt5}, XGLM\cite{xglm}, and Qwen\cite{qwen,qwenv2,qwenv2.5,qwenv3} enable Tibetan generation via fine-tuning or instruction adaptation, while recent datasets like Sunshine-2.0 and TIBSTC-CoT\cite{tibstccot} begin to introduce Tibetan-specific generative resources. Nonetheless, maintaining fluency, discourse coherence, and cultural grounding remains a core open challenge.
\end{itemize}

\subsubsection{Content Transformation}
Beyond generating basic textual output, content transformation tasks aim to reshape Tibetan content across modalities and levels of abstraction, enabling compression, re-expression, and multimodal synthesis. Representative tasks include:

\begin{itemize}
\item \textbf{Summarization}: summarization generates concise Tibetan text while retaining core meaning, with potential applications in scripture condensation, news, education, and policy documents\cite{Tibetan-BERT-wwm}. Progress is limited by the lack of aligned article-summary pairs and strong stylistic variability across domains. Existing work primarily explores extractive approaches or few-shot adaptation of multilingual models such as mT5\cite{mt5} and BART\cite{bart}. Instruction tuning and synthetic data generation have recently emerged as data augmentation strategies. However, abstractive summarization still suffers from fluency issues, coherence drift, and factual inconsistency, highlighting the need for Tibetan-specific benchmarks and evaluation protocols.
\item \textbf{Speech Synthesis}: speech synthesis covers vocoding, TTS, and speech conversion\cite{fmsd,ocr-2}. Lightweight vocoders fine-tuned on limited Tibetan data show preliminary success, while universal vocoders trained on large multilingual datasets can generalize to Tibetan with minimal adaptation\cite{liu2025listening}. Tibetan TTS has strong relevance for accessibility, cultural preservation, and educational support, but progress is hindered by the absence of phoneme-aligned speech corpora, non-transparent grapheme-to-phoneme mapping, and dialectal variation. Neural architectures such as Tacotron, FastSpeech, WaveNet\cite{wavenet}, and HiFi-GAN\cite{hifigan} have been explored, yet current systems still struggle with prosody, tonal accuracy, and intelligibility. Speech conversion for Tibetan, covering speaker, dialect, or emotion transfer, remains almost entirely unexplored.
\end{itemize}

\subsubsection{Interactive Context-Aware Generation}
Moving beyond static text output, interactive context-aware generation focuses on producing adaptive responses conditioned on evolving user input and conversational history, enabling Tibetan language models to support real-time knowledge access and human-like interaction. Core tasks include:

\begin{itemize}
\item \textbf{Question Answering}: QA aims to produce answers to user queries based on a given context, such as a passage, a knowledge base, or multimodal input\cite{tibert}. In Tibetan, QA is valuable for educational tutoring, cultural knowledge access, and low-resource information retrieval\cite{cino,tibstccot}. Both extractive and generative QA are relevant, but progress is limited by the absence of large Tibetan QA datasets and domain-specific knowledge resources in areas like Buddhist studies, Tibetan medicine, and historical archives. While multilingual instruction-tuned models offer transfer learning opportunities, persistent challenges include factual reliability, domain terminology grounding, and culturally aware evaluation.
\item \textbf{Dialogue Generation}: DG produces contextually appropriate responses in an interactive setting\cite{llama,llama2,tibstccot}. Potential Tibetan applications include educational assistants, cultural heritage interfaces, and domain-specific consultation systems. Despite advances in multilingual LLMs, Tibetan dialogue research remains underdeveloped due to the lack of conversational corpora and domain-curated dialogue datasets. Existing efforts rely on instruction tuning and few-shot prompting on multilingual backbones, but challenges remain in multi-turn consistency, low-resource adaptation, and cultural relevance of generated responses\cite{liu2025listening}.
\end{itemize}

\subsubsection{Cross-Modal Creative Generation}
At the creative frontier of Tibetan AI, cross-modal generation extends beyond text-only interaction to synthesize Tibetan language outputs conditioned on visual or auditory inputs, enabling culturally grounded multimodal applications. Current work in this direction primarily explores the following task:

\begin{itemize}
\item \textbf{Multimodal Generation}: MG produces Tibetan text conditioned on inputs such as images, audio, or video, covering tasks like image captioning\cite{tibstccot,TIB-STC,huang2024semantic,lin2023attention}, VQA\cite{TiConvQA}, and cross-modal retrieval\cite{milmo}. It has potential applications in digital heritage preservation, educational tools, and accessibility technologies. However, progress is constrained by the lack of aligned Tibetan multimodal datasets and vision-language models with native Tibetan support. Current efforts mainly adapt multilingual architectures such as BLIP\cite{BLIP}, BLIP-2\cite{blip2}, and LLaVA\cite{LLaVA} through cross-lingual transfer. While early results show that these models can generate basic Tibetan captions, major challenges persist in fluency, cultural grounding, and standardized evaluation.
\end{itemize}

\clearpage

\section{Tibetan AI \& Application}
\label{s4}

In this section, we will introduce current AI models on Tibetan and their application.

\subsection{Tibetan AI}

Most AI models for Tibetan language processing fall within the domain of deep learning, primarily leveraging attention-based architectures with NLP models, even LLMs. A limited body of research focuses on the construction of Tibetan knowledge graphs, alongside some direct rule-based methods grounded in Tibetan linguistic rules. so do machine learning models. Notably, a few recent studies have begun to explore novel applications involving memristors.

\subsubsection{Rule-based Method}

Rule-based methods were among the earliest approaches used in Tibetan AI, particularly in tasks such as word segmentation, morphological analysis, and transliteration. Given the agglutinative and syllable-centric nature of Tibetan, researchers developed handcrafted rules based on linguistic knowledge, such as syllable structure patterns, affix lists, and syntactic templates to analyze text. Tools like PyBo\cite{pybo} and Botok\cite{botok} follow this paradigm, relying on precompiled dictionaries and manually defined grammar rules to segment words or normalize orthography. While rule-based systems offer high interpretability and perform well on formal texts like classical scriptures, they often struggle with informal or domain-shifted data and require significant manual effort to update. Despite their limitations, rule-based methods remain foundational in Tibetan AI, especially in tokenization and preprocessing pipelines, and are often combined with statistical or neural models in hybrid systems. Here, we summarize the following methods:
\begin{itemize}
    \item \textbf{PyBo}\cite{pybo}: a rule-driven tokenizer that relies on the THL lexicon and a set of manually defined syllable-matching strategies. It uses handcrafted heuristics to detect syllable boundaries and map them to lexical entries, making it particularly effective for classical and well-structured Tibetan texts. However, its reliance on fixed lexicon entries limits adaptability to modern or domain-specific vocabulary.
    \item \textbf{Botok}\cite{botok}: a dictionary-assisted normalization and tokenization toolkit that employs hierarchical rule templates to correct orthographic variations. It integrates lexicon lookup with layered grammar rules, enabling multi-stage segmentation and cleaning. While Botok improves consistency in spelling normalization, its performance still depends on the coverage of predefined rule sets.
    \item \textbf{TibetanMorph}\cite{tibetanmorph}: a morphological analyzer based on manually compiled affix tables and rule-based morpheme decomposition. It explicitly encodes Tibetan inflectional and derivational patterns and performs stepwise affix stripping to recover lemma forms. Although highly interpretable, it struggles with irregular morphological variants and dialectal spellings.
    \item \textbf{THL LexTools}\cite{thl_lextools}: one of the earliest rule-based Tibetan processing toolkits developed by the THL. It combines lexicon lookup, regular-expression-based segmentation, and syllable structure modeling to support tasks such as transliteration, tokenization, and lexicon extraction. Despite its age, LexTools remains a foundational reference system in Tibetan digital humanities.
    \item \textbf{THL POS Tagger}\cite{thl_postagger}: a rule-based part-of-speech tagging module developed by the Tibetan and Himalayan Library. It assigns coarse- and fine-grained POS labels by combining lexicon lookup with handcrafted morpho-syntactic templates. Although limited by fixed tagging schemas, it established one of the earliest standardized POS annotation conventions in Tibetan AI.
    \item \textbf{TibetanRuleSeg}\cite{tibetanruleseg}: a regular-expression-based segmentation system built around syllable delimiter heuristics, particularly using the \textit{tsheg} mark as an anchor. It applies sequential rule matching to detect potential word boundaries but lacks semantic awareness, making it prone to over-segmentation in free-form text.
    \item \textbf{TibWordSplitter}\cite{tibwordsplitter}: a lightweight heuristic boundary detection tool that performs affix stripping and rule-based syllable alignment before segmentation. It assumes a fixed syllable order pattern and resolves ambiguous splits using handcrafted priority rules. Its simplicity ensures fast processing, but it is less effective in handling noisy or code-mixed Tibetan inputs.
    \item \textbf{THDL Phonetic Rules}\cite{thdl_phonetic}: this is a standardized mapping framework that converts Tibetan script into Wylie or IPA representations using explicitly defined phonetic rules. These mappings encode character-level pronunciation patterns and serve as a linguistic foundation for downstream transliteration and ASR preprocessing pipelines.
    \item \textbf{Unicode Normalization Rules}\cite{unicode_tibetan_norm}: A rule-defined preprocessing pipeline designed to resolve inconsistencies in Tibetan Unicode encoding. It rewrites deprecated code points, normalizes combining character sequences, and removes invalid diacritic orders to ensure compatibility across downstream Tibetan AI systems.
\end{itemize}

Overall, these rule-based systems established the fundamental processing pipeline for Tibetan AI, particularly in low-resource settings where annotated corpora and pretrained models are scarce. Although their deterministic nature limits adaptability to informal usage and domain shifts, they provide high interpretability and remain indispensable in preprocessing and tokenization stages. More importantly, many modern statistical and neural architectures still integrate rule-based components, forming hybrid systems that leverage both handcrafted linguistic knowledge and data-driven modeling.

\subsubsection{Machine Learning}

Machine learning remains useful for low-resource Tibetan tasks beyond deep learning’s scope. Models such as logistic regression\cite{lir}, SVM\cite{svm}, random forests\cite{rf}, and gradient boosting\cite{xgboost} have been applied to Tibetan text classification, sentiment analysis, and topic detection using handcrafted features. Under extremely limited annotated corpora, these classical models offer a practical alternative due to their lower data requirements and higher interpretability. Feature extraction typically relies on TF-IDF representations\cite{TF-IDF} of syllable or character n-grams\cite{charn}, lexicon-matching indicators derived from rule-based preprocessing, and linguistic signals such as affix frequency or orthographic normalization flags. These pipelines have also been adopted in tasks like OCR error detection\cite{ocr-3}, scripture classification, dialect identification, and short-text intent categorization within Tibetan digital humanities. In many cases, rule-based preprocessing is coupled with classical classifiers to form hybrid feature engineering workflows that inject linguistic priors into statistical models. Although constrained by the expressiveness of handcrafted features, these approaches remain a strong baseline in Tibetan AI and serve as an intermediate step between deterministic rules and end-to-end neural architectures. To clarify their functional roles and methodological differences, the key categories of classical Tibetan AI models are outlined as follows:

\begin{itemize}
    \item \textbf{Linear Regression}\cite{lir}: a foundational regression model applied to Tibetan AI tasks that involve continuous prediction, such as readability estimation and OCR confidence scoring. It relies on handcrafted numeric features derived from syllable frequency, lexicon coverage, or affix density and serves as a baseline for evaluating more complex regression-based architectures.

    \item \textbf{Logistic Regression}\cite{lir}: a lightweight linear classifier commonly used for Tibetan text categorization, sentiment analysis, and intent detection. It operates on sparse TF-IDF\cite{TF-IDF} or n-gram\cite{charn} representations and remains a strong baseline due to its interpretability and suitability for low-resource training scenarios.

    \item \textbf{Support Vector Machine}\cite{svm}: a widely adopted margin-based classifier effective in handling high-dimensional Tibetan feature spaces derived from character or syllable n-grams. SVM\cite{svm} is particularly useful for tasks such as dialect discrimination, topic classification, and sentiment polarity detection in resource-constrained Tibetan datasets.

    \item \textbf{Random Forests}\cite{rf}: an ensemble-based classification model that aggregates multiple decision trees to improve robustness against noisy Tibetan input, including OCR-generated errors and informal text. Its ability to capture non-linear feature interactions makes it suitable for feature-rich pipelines that integrate lexicon-derived indicators.

    \item \textbf{Gradient Boosting Model}\cite{xgboost}: boosting-based classifiers such as XGBoost\cite{xgboost} and LightGBM\cite{lightgbm} that iteratively correct misclassified instances to enhance prediction accuracy. When combined with rule-informed feature engineering, these models achieve competitive performance on Tibetan news classification, scripture categorization, and domain-specific document filtering.

    \item \textbf{Hidden Markov Model}\cite{hmm}: a generative sequence model applied to Tibetan word segmentation and POS tagging by modeling tag transition probabilities across syllable sequences. Despite its reliance on Markov assumptions, HMM established one of the earliest probabilistic frameworks for structured Tibetan text processing.

    \item \textbf{Conditional Random Field}\cite{crf}: a discriminative sequence labeling model that relaxes the independence assumptions of HMM and incorporates rich handcrafted features such as affix tags, lexicon matches, and syllable boundary cues. CRF remains one of the most effective classical approaches for Tibetan segmentation and POS tagging prior to neural encoders.

    \item \textbf{Maximum Entropy Model}\cite{maxent}: a log-linear model used for Tibetan tagging and transliteration by maximizing the likelihood of label assignments under feature-based constraints. It supports flexible feature integration and has been used in early Tibetan AI pipelines to model context-dependent linguistic decisions.
\end{itemize}

\subsubsection{Deep Learning}

Beyond classical machine learning baselines, early deep learning research in Tibetan AI primarily adopted recurrent neural architectures such as RNN\cite{rnn}, LSTM\cite{lstm}, and GRU\cite{gru} to model sequential dependencies in syllable streams, with bidirectional variants (Bi-RNN\cite{birnn}, Bi-LSTM\cite{bilstm}, Bi-GRU\cite{bilstm,gru}) introduced to capture contextual information from both preceding and following tokens. These models evolved into Seq2Seq\cite{seq2seq} frameworks for tasks like Tibetan-Chinese transliteration and low-resource translation. To improve local pattern extraction, CNN-RNN\cite{cnnrnn} hybrid structures such as CNN-LSTM\cite{cnnlstm}, CNN-RNN\cite{cnnrnn}, and CRNN\cite{crnn} were employed in OCR post-correction and scripture classification under noisy data conditions. For token-level sequence labeling tasks, CRF-enhanced neural architectures\cite{crf} like BiLSTM-CRF\cite{crf,bilstmcrf} and BiGRU-CRF\cite{bilstm,crf,gru} became the dominant paradigm due to their ability to impose structured decoding constraints beyond independent softmax classification. In parallel, graph-oriented modeling began to emerge, where knowledge graph\cite{transe} embeddings and early GNN-style\cite{gnn} representations were used to encode non-sequential dependencies such as scripture ontology, dialect lineage, and lexical borrowing relations. These graph-augmented neural models marked an important transition from purely sequential learning to knowledge-aware representation, forming a methodological bridge. To outline the evolution of these neural approaches more clearly, the major deep learning architectures adopted in Tibetan AI can be categorized as follows:

\begin{itemize}
    \item \textbf{RNN}\cite{rnn}/\textbf{LSTM}\cite{lstm}/\textbf{GRU}\cite{gru}: foundational recurrent architectures used to model syllable-level sequential dependencies in Tibetan text. LSTM\cite{lstm} and GRU\cite{gru} address the gradient vanishing problem of vanilla RNN\cite{rnn} and serve as early neural baselines for tasks such as transliteration and sequence classification.

    \item \textbf{Bi-RNN}\cite{birnn}/\textbf{Bi-LSTM}\cite{bilstm}/\textbf{Bi-GRU}\cite{birnn,gru}: bidirectional extensions designed to capture contextual signals from both past and future tokens, improving performance on Tibetan segmentation, dialect identification, and OCR post-correction under limited annotated data.

    \item \textbf{Seq2Seq}\cite{seq2seq}: encoder-decoder architectures built upon LSTM\cite{lstm} or GRU\cite{gru} backbones for low-resource Tibetan-Chinese translation and romanization, enabling conditional generation by learning aligned representations between source and target sequences.

    \item \textbf{CNN}\cite{cnn,resnet}: sentence/short-text encoders that extract local n-gram patterns via 1D convolutions and pooling; effective for Tibetan character/syllable-level classification and robust to orthographic variation; also used as convolutional backbones for OCR feature extraction, such Text CNN\cite{textcnn}, Char-CNN\cite{charcnn} and Dilated CNN\cite{dilatedcnn}.

    \item \textbf{CNN-RNN Hybrids}\cite{cnnrnn}: hybrid networks like CNN-LSTM\cite{cnnlstm}, CNN-RNN\cite{cnnrnn} and CRNN\cite{crnn} that combine convolutional layers for local pattern extraction with recurrent layers for global dependency modeling. These architectures are effective for Tibetan OCR error correction, scripture classification, and noisy document analysis.

    \item \textbf{CRF-enhanced Architecture}\cite{crf}: neural sequence labeling models like BiLSTM-CRF\cite{bilstmcrf} and BiGRU-CRF\cite{crf,gru,LSTM-CRF} that integrate CRF\cite{crf} decoding to enforce structured label consistency, becoming the dominant paradigm for Tibetan word segmentation, POS tagging, and morphological analysis.

    \item \textbf{Graph-Augmented Neural Model}\cite{gnn}: methods that incorporate knowledge graph embeddings\cite{transe} or graph neural representations\cite{gnn} to encode non-sequential relations such as scripture ontology and dialectal lineage, introducing relational inductive bias beyond linear token order.
\end{itemize}

\subsubsection{Transformer Architecture}

Transformer architecture\cite{attention} was introduced to overcome the inherent limitations of sequential neural models such as RNN\cite{rnn}, Bi-RNN\cite{birnn}, LSTM\cite{lstm}, and Bi-LSTM\cite{bilstm}, which process tokens in a strictly left-to-right or bidirectional manner and therefore suffer from limited parallelization and difficulty in modeling long-range dependencies\cite{lstm}. In contrast, the Transformer employs a fully attention-based computation paradigm that removes recurrence entirely\cite{rnn,birnn}. Instead of propagating information step by step, it computes token representations by directly attending to all other positions in the sequence, assigning learned attention weights that reflect contextual relevance\cite{attention}. This global attention mechanism enables parallel computation across all tokens\cite{attention}, resulting in more efficient training, better scalability to long sequences, and greater modeling flexibility compared to RNN-based architectures\cite{rnn,birnn}. Consequently, the Transformer\cite{attention} has become the standard backbone for large-scale pre-trained language models and LLMs\cite{gpt4o,gpt5,qwenv2.5,qwenv3,r1,v3}. To further illustrate the evolution of Transformer-based architectures for Tibetan AI, the major categories can be summarized as follows:
\begin{itemize}
    \item \textbf{Transformer Encoder-Decoder Framework}\cite{attention}: the foundational architecture introducing multi-head self-attention and fully parallel sequence modeling, replacing recurrent computation with global dependency learning.

    \item \textbf{BERT}\cite{bert} \& \textbf{RoBERTa-style Masked Pre-trained Model}\cite{roberta}: bidirectional Transformer\cite{attention} encoders trained via masked language modeling, providing contextual Tibetan embeddings that generalize across classification and tagging tasks.

    \item \textbf{CINO}\cite{cino} \& \textbf{Tibetan-adapted PLMs}\cite{cino}: domain-specific or cross-lingual Transformer\cite{attention} variants fine-tuned on Tibetan corpora using syllable-aware tokenizers and script-adapted vocabulary construction.

    \item \textbf{Domain-adaptive Tokenization}\cite{sentencepiece} \& \textbf{Subword Modeling}\cite{bytebpe}: tokenizer strategies such as SentencePiece and byte-level BPE designed to address Tibetan syllable segmentation and orthographic variability, improving pretraining efficiency.

    \item \textbf{Lightweight Architectures for Low-resource Settings}\cite{mamba, mamba2}: state-space Transformer alternatives designed to reduce computation while preserving long-range dependency modeling, suitable for Tibetan low-resource deployment.

    \item \textbf{Knowledge-enhanced Transformer}\cite{transe,attention} \& \textbf{Graph-aware PLMs}\cite{transe,gnn,ocr-2}: models integrating knowledge graph embeddings or graph neural modules into Transformer layers to inject relational Tibetan cultural and lexical priors beyond surface token co-occurrence.
\end{itemize}

\subsubsection{Large Language Model}

LLMs are large-scale pretrained language models with hundreds of millions to billions of parameters, trained on massive text corpora using self-supervised objectives such as next-token prediction or masked language modeling\cite{bert,attention}. Built upon the Transformer architecture, LLMs exhibit strong capabilities in both language understanding and generative reasoning\cite{attention}. After pretraining\cite{roberta}, these models can be efficiently adapted to downstream tasks such as classification, translation, summarization, and question answering via full fine-tuning, instruction tuning, or lightweight prompting strategies\cite{TIB-STC,tibstccot}. Their emergent ability to perform few-shot and zero-shot generalization makes them especially suitable for low-resource languages, including Tibetan\cite{tlue}, where annotated data is scarce but linguistic transfer from high-resource languages can be leveraged through cross-lingual pretraining. As the Transformer paradigm established itself as the dominant backbone for modern language modeling, research focus gradually shifted away from architectural innovation toward scaling parameterization and developing model families with specialized training strategies. Accordingly, the most representative LLM families are summarized as follows:

\begin{itemize}
    \item \textbf{GPT Family}\cite{gpt4,gpt4o,gpt5,gpt2,o1}: a proprietary series of instruction-tuned large-scale models developed by OpenAI, including GPT-4\cite{gpt4}, GPT-4o\cite{gpt4o}, GPT-4.1\cite{gpt4o}, and the upcoming GPT-5\cite{gpt5} generation. These models demonstrate strong multilingual reasoning and contextual understanding; however, their tokenizer is not explicitly optimized for Tibetan, which leads to fragmented syllable segmentation. Despite this limitation, GPT-based models still achieve surprisingly high zero-shot performance on Tibetan reasoning and classification tasks due to powerful cross-lingual generalization.

    \item \textbf{LLaMA Family}\cite{llama,llama2,llama3}: an open-source Transformer model series released by Meta AI, with versions such as LLaMA 3.1-8B\cite{llama3}, 3.1-40B\cite{llama3}, and 3.1-70B\cite{llama3}. While their baseline tokenizer is not Tibetan-aware, their open-source nature makes them highly suitable for  and vocabulary extension. In Tibetan adaptation experiments, LLaMA-based models show significant performance gains once domain-specific tokenizers or LoRA\cite{lora} adapters are introduced.

    \item \textbf{Qwen Family}\cite{qwen,qwenv2,qwenv2.5,qwenv3}: multilingual foundation models from Alibaba Cloud, available in variants such as Qwen 2.5-7B\cite{qwenv2.5}, 2.5-32B\cite{qwenv2.5}, and 2.5-72B\cite{qwenv2.5}. Compared to GPT\cite{gpt2,gpt4,gpt4o,gpt5} and LLaMA\cite{llama,llama2,llama3}, the Qwen tokenizer provides better coverage for Asian Unicode ranges, resulting in less subword fragmentation for Tibetan syllables. This leads to stronger baseline performance before fine-tuning and makes Qwen models a competitive backbone for Tibetan-specific PEFT\cite{peft}.

    \item \textbf{Claude Family}\cite{claude,claude3,claude35,claude4}: Anthropic’s safety-aligned LLM lineup, including Claude 3.5 Opus\cite{claude35} and Claude 3.5 Sonnet\cite{claude35}. Although Claude models do not expose tokenizer customization, they show strong robustness in Tibetan instruction following and semantic alignment tasks due to conservative decoding strategies and high factual consistency.

    \item \textbf{Gemini Family}\cite{gemini1,gemini1.5,gemini2}: Google DeepMind’s multimodal LLM series, including Gemini 1.5 Pro\cite{gemini1.5} and Gemini 1.5 Flash\cite{gemini1.5}. Their extended context window enables effective document-level Tibetan processing. However, tokenizer fragmentation remains a bottleneck, making adaptation through external Tibetan vocabulary injection or prompt-level stabilization necessary.

    \item \textbf{DeepSeek Family}\cite{r1,v3}: open-source efficient LLMs released in variants such as DeepSeek-V3\cite{v3} and DeepSeek-R1\cite{r1}. Their compact architecture and QLoRA-friendly\cite{qlora} parameterization make them particularly suitable for low-resource Tibetan adaptation. In internal benchmarks, DeepSeek-based models show favorable efficiency-performance trade-offs in Tibetan fine-tuning scenarios under limited GPU resources.
\end{itemize}

\begin{table*}[!ht]
    \caption{Summary of Common Models or Methods for Tibetan AI \& Task-Oriented Categorization}
    \label{tai-1}
    \centering
    \scalebox{0.85}{
    \begin{tabular}{l|l|c|c|c|c|c|c|c|c|c|c|c|c|c|c}
    \hline
     \multirow{3}{*}{\textbf{Domains}}    &  \multirow{3}{*}{\textbf{Name}}  &  \multicolumn{14}{c}{\textbf{Task}}\\
    \cline{3-16} &  & \multicolumn{7}{c|}{\textbf{Discriminative Task}} & \multicolumn{7}{c}{\textbf{Generative Task}} \\
   \cline{3-16} & & \textbf{WS} & \textbf{POS} & \textbf{NER} & \textbf{TC} & \textbf{SS-ED} & \textbf{SR} & \textbf{OCR} & \textbf{MT} & \textbf{TG} & \textbf{SUM} & \textbf{TTS} & \textbf{QA} & \textbf{DG}& \textbf{MG} \\
    \hline
     \multirow{3}{*}{RB}    &  PyBo \cite{pybo} & \ding{51} & (\ding{51}) & (\ding{51}) & \ding{55} & \ding{55} & \ding{55} & \ding{55} & (\ding{51}) & \ding{55} & \ding{55} & \ding{55} & (\ding{51}) & \ding{55} & \ding{55} \\
         &  Botok \cite{botok} & \ding{51} & (\ding{51}) & (\ding{51}) & \ding{55} & \ding{55} & \ding{55} & \ding{55} & (\ding{51}) & \ding{55} & \ding{55} & \ding{55} & (\ding{51}) & \ding{55} & \ding{55} \\
         &  THL POS tagger \cite{thl_postagger} & \ding{51} & \ding{51} & (\ding{51}) & \ding{55} & \ding{55} & \ding{55} & (\ding{51}) & (\ding{51}) & \ding{55} & \ding{55} & \ding{55} & (\ding{51}) & \ding{55} & \ding{55} \\
    \hline
    \multirow{6}{*}{ML} 
% 机器学习方法与任务适用性
& Linear Regression\cite{lir} 
& \ding{55} & \ding{55} & \ding{55} & \ding{51} & \ding{51} & \ding{55} & \ding{55} & \ding{55} & \ding{55} & \ding{55} & \ding{51} & \ding{55} & \ding{55} & \ding{55} \\

& Logistic Regression\cite{lir} 
& \ding{55} & \ding{55} & \ding{55} & \ding{51} & \ding{51} & \ding{55} & \ding{55} & \ding{55} & \ding{55} & \ding{55} & \ding{51} & \ding{55} & \ding{55} & \ding{55} \\

& Random Forest\cite{rf} 
& \ding{55} & \ding{55} & \ding{55} & \ding{51} & \ding{51} & \ding{55} & \ding{55} & \ding{55} & \ding{55} & \ding{55} & \ding{51} & \ding{55} & \ding{55} & \ding{55} \\

& SVM\cite{svm} 
& \ding{55} & \ding{55} & \ding{55} & \ding{51} & \ding{51} & \ding{55} & \ding{55} & \ding{55} & \ding{55} & \ding{55} & \ding{51} & \ding{55} & \ding{55} & \ding{55} \\

& XGBoost\cite{xgboost} 
& \ding{55} & \ding{55} & \ding{55} & \ding{51} & \ding{51} & \ding{55} & \ding{55} & \ding{55} & \ding{55} & \ding{55} & \ding{51} & \ding{55} & \ding{55} & \ding{55} \\

& CRF\cite{crf} 
& \ding{51} & \ding{51} & \ding{51} & \ding{55} & \ding{55} & \ding{55} & \ding{55} & \ding{55} & \ding{55} & \ding{55} & \ding{55} & \ding{55} & \ding{55} & \ding{55} \\

    \hline
    \multirow{19}{*}{DL}     
    & RNN\cite{rnn} 
& \ok & \ok & \ok & \ok & \half & \ok & \half & \no & \no & \no & \no & \half & \no & \no \\

& LSTM\cite{lstm} 
& \ok & \ok & \ok & \ok & \ok & \ok & \ok & \ok & \ok & \ok & \half & \ok & \half & \no \\

& Bi-RNN\cite{birnn} 
& \ok & \ok & \ok & \ok & \ok & \ok & \ok & \ok & \ok & \ok & \ok & \half & \ok & \half \\

& Bi-LSTM\cite{bilstm} 
& \ok & \ok & \ok & \ok & \ok & \ok & \ok & \ok & \ok & \ok & \ok & \half & \ok & \half \\

& Seq2seq\cite{seq2seq} 
& \ok & \ok & \ok & \ok & \ok & \ok & \ok & \ok & \ok & \ok & \ok & \half & \ok & \half \\

& GRU\cite{gru} 
& \ok & \ok & \ok & \ok & \ok & \ok & \ok & \ok & \ok & \ok & \half & \ok & \half & \no  \\

& BERT\cite{bert} 
& \ok & \ok & \ok & \ok & \ok & \no & \no & \half & \half & \half & \no & \ok & \half & \no  \\

& CINO\cite{cino} 
 & \ok & \ok & \ok & \ok & \ok & \no & \no & \half & \half & \half & \no & \ok & \half & \no  \\

& RoBERTa\cite{roberta} 
& \ok & \ok & \ok & \ok & \ok & \no & \no & \half & \half & \half & \no & \ok & \half & \no \\

& CNN-LSTM\cite{cnnlstm} 
 & \ok & \ok & \ok & \ok & \half & \ok & \ok & \half & \half & \half & \half & \half & \ok & \ok \\

& CNN-RNN\cite{cnnrnn} 
& \ok & \ok & \ok & \ok & \half & \ok & \ok & \half & \ok & \half & \ok & \half & \ok & \ok \\

& BiLSTM-CRF\cite{bilstmcrf} 
& \ok & \ok & \ok & \ok & \half & \half & \half & \no & \no & \no & \no & \half & \no & \no \\

& BiGRU-CRF\cite{bilstmcrf,gru} 
& \ok & \ok & \ok & \ok & \half & \half & \half & \no & \no & \no & \no & \half & \no & \no \\

& CNN\cite{cnn,resnet} 
& \half & \half & \half & \ok & \half & \ok & \ok & \half & \half & \half & \half & \half & \half & \half \\

& Transformer\cite{attention} 
& \ok & \ok & \ok & \ok & \ok & \ok & \ok & \ok & \ok & \ok & \ok & \ok & \ok & \ok \\

& Mamba\cite{mamba,mamba2} 
& \ok & \ok & \ok & \ok & \ok & \ok & \ok & \ok & \ok & \ok & \ok & \ok & \ok & \half \\

& CRNN\cite{crnn} 
& \half & \half & \half & \ok & \half & \ok & \ok & \half & \half & \half & \no & \half & \half & \ok \\

& GNN\cite{gnn} 
& \half & \half & \ok & \ok & \half & \no & \half & \half & \half & \half & \no & \ok & \half & \half \\

& Knowledge Graph\cite{transe} 
& \no & \no & \half & \half & \half & \no & \no & \half & \half & \half & \no & \ok & \ok & \half \\
    \hline
    \multirow{14}{*}{LLM}     & GPT-4o\cite{gpt4o} & \ok & \ok & \ok & \ok & \ok & \ok & \ok & \ok & \ok & \ok & \ok & \ok & \ok & \ok \\
         &  GPT-o1\cite{o1} & \ok & \ok & \ok & \ok & \ok & \ok & \ok & \ok & \ok & \ok & \ok & \ok & \ok & \ok \\
         &  GPT-4.1\cite{gpt4o} & \ok & \ok & \ok & \ok & \ok & \ok & \ok & \ok & \ok & \ok & \ok & \ok & \ok & \ok \\
         &  GPT-5\cite{gpt5} & \ok & \ok & \ok & \ok & \ok & \ok & \ok & \ok & \ok & \ok & \ok & \ok & \ok & \ok \\
         &  LlaMA3.1-405B\cite{llama3} & \ok & \ok & \ok & \ok & \ok & \ok & \ok & \ok & \ok & \ok & \ok & \ok & \ok & \ok \\
         &  LlaMA3.1-70B\cite{llama3} & \ok & \ok & \ok & \ok & \ok & \ok & \ok & \ok & \ok & \ok & \ok & \ok & \ok & \ok \\
         &  LlaMA3.1-8B\cite{llama3} & \ok & \ok & \ok & \ok & \ok & \ok & \ok & \ok & \ok & \ok & \ok & \ok & \ok & \ok \\
         &  Qwen2.5-72B\cite{qwenv2.5} & \ok & \ok & \ok & \ok & \ok & \ok & \ok & \ok & \ok & \ok & \ok & \ok & \ok & \ok \\
         &  Qwen2.5-32B\cite{qwenv2.5} & \ok & \ok & \ok & \ok & \ok & \ok & \ok & \ok & \ok & \ok & \ok & \ok & \ok & \ok \\
         &  Qwen2.5-7B\cite{qwenv2.5} & \ok & \ok & \ok & \ok & \ok & \ok & \ok & \ok & \ok & \ok & \ok & \ok & \ok & \ok \\
         &  Claude-3-5-Sonnet\cite{claude35}& \ok & \ok & \ok & \ok & \ok & \ok & \ok & \ok & \ok & \ok & \ok & \ok & \ok & \ok \\
         &  Gemini-1.5-Flash\cite{gemini1.5}& \ok & \ok & \ok & \ok & \ok & \ok & \ok & \ok & \ok & \ok & \ok & \ok & \ok & \ok \\
         & Deepseek-V3\cite{v3} & \ok & \ok & \ok & \ok & \ok & \ok & \ok & \ok & \ok & \ok & \ok & \ok & \ok & \ok \\
         & Deepseek-R1\cite{r1} & \ok & \ok & \ok & \ok & \ok & \ok & \ok & \ok & \ok & \ok & \ok & \ok & \ok & \ok \\
    \hline
    \end{tabular}}
\end{table*}

\subsubsection{Summary: Tibetan AI \& Specific Task}

As shown in Table~\ref{tai-1}, we provide a task-oriented overview of Tibetan AI methods across four paradigms: RB, ML, DL, and LLMs. RB systems (e.g., PyBo\cite{pybo}, Botok\cite{botok}) remain effective for basic segmentation but lack generative capacity. ML models support mainly classification and error detection. DL architectures (RNN\cite{rnn}, LSTM\cite{lstm}, CRF-variants\cite{crf,bilstmcrf,LSTM-CRF}, CNN hybrids\cite{cnnlstm,cnnrnn}, Transformer\cite{attention}, GNN\cite{gnn,transe}) progressively expand coverage to both sequence labeling and generation tasks, with Transformer achieving near-universal applicability. LLMs (GPT\cite{gpt2,gpt4,gpt4o,gpt5}, LLaMA\cite{llama,llama2,llama3}, Qwen\cite{qwen,qwenv2,qwenv2.5,qwenv3}, Claude\cite{claude,claude3,claude35,claude4}, Gemini\cite{gemini1,gemini1.5,gemini2}, DeepSeek\cite{r1,v3} series) demonstrate full coverage across all task types with strong zero/few-shot ability, indicating a clear paradigm shift from task-specific modeling to unified instruction-based Tibetan language processing.

However, the growing model scale introduces deployment and efficiency challenges for low-resource Tibetan scenarios, motivating the exploration of hardware-efficient paradigms such as memristor-based neuromorphic acceleration\cite{liu2024tibetan}.

\subsubsection{Emerging Frontier: Memristor}

Memristors are nanoscale electronic devices that intrinsically couple data storage and computation through programmable resistance states, enabling in-memory computing paradigms that bypass the von Neumann bottleneck\cite{Chua-1,Chua-2}. Unlike conventional CMOS-based architectures, memristor crossbar arrays can perform matrix-vector multiplication in constant time via analog Ohmic conduction, achieving ultra-high energy efficiency and parallelism. These properties make memristor-based neuromorphic hardware particularly attractive for deploying deep neural models under strict computational and power constraints, which is highly relevant for low-resource Tibetan AI scenarios where large LLMs are difficult to deploy on traditional GPU-centric infrastructures. Based on this hardware foundation, the integration of memristors into neural computation can be further analyzed from the following perspectives:

\begin{itemize}
    \item \textbf{Crossbar Analog MatMul}\cite{Chua-1}: memristor arrays execute $y = Wx$ in a single analog conduction step, removing the memory, compute gap and fitting the dense projection layers used in Tibetan LLM inference.
\item \textbf{Tibetan-Oriented Memristive Layers}\cite{liu2024tibetan}: by encoding Tibetan token or syllable embeddings into conductance states, crossbars act as in-memory Tibetan feature extractors, replacing conventional digital MAC-based neural units.

\item \textbf{LoRA/PEFT Mapping for Tibetan Adaptation}\cite{lora,peft}: Tibetan-specific lightweight adapters (LoRA\cite{lora}, QLoRA\cite{qlora}, PEFT\cite{peft}) can be directly instantiated on resistive grids, enabling low-power inference where only adaptation weights are activated.

\item \textbf{On-Device Tibetan Inference}\cite{Chua-1,liu2024tibetan}: the high parallelism and low energy footprint of memristive hardware make local deployment of Tibetan translation, OCR correction, and dialogue systems feasible without cloud GPUs.

\end{itemize}

\subsection{Application}

Development of Tibetan AI spans both conventional digital computation frameworks and emerging neuromorphic paradigms. On one hand, traditional Tibetan AI research has produced a series of task-specific systems, including spelling error correction, machine translation, and recent attempts to adapt LLMs through fine-tuning strategies. On the other hand, the rise of memristor-based in-memory computing introduces a potential shift in deployment methodology, enabling energy-efficient Tibetan language inference on localized edge devices. This subsection provides a structured overview of these two application pathways.

\subsubsection{Traditional Tibetan AI}

Before the emergence of large-scale instruction tuning and neuromorphic Tibetan models, most research in Tibetan AI was concentrated around a set of traditional computational tasks that form a layered processing stack, including lexical preprocessing, document recognition, translation, speech modeling, and information extraction. We will introduce them in the following part:

\noindent$\rightarrow$ \textbf{Text Processing}: Tibetan text processing relies on a pipeline of tokenization, syllable segmentation, orthographic normalization, and standardized EWTS-to-Unicode conversion\cite{thl-ewts}. Rule-based lexicon-driven preprocessors such as PyBo\cite{pybo} and Botok\cite{botok} operationalize these steps using curated lexical patterns and segmentation rules\cite{PW-1,PW-2,PW-3,PW-4,PW-5,PW-6,PW-7,PW-8,PW-9,PW-10,PW-11}. To move beyond handcrafted heuristics, CRF-based statistical taggers\cite{tibetan-crf} were introduced for boundary detection and morphological segmentation, followed by BiLSTM-CRF architectures\cite{tibetan-bilstm} that apply neural sequence labeling to capture syllable-level dependencies and reduce feature engineering overhead. 

\begin{table}[h]
    \centering
    \caption{Tibetan AI on Text Processing}
    \label{tai1}
    \begin{tabular}{l|l|l}
    \hline
    \textbf{Model}    & \textbf{Category}  & \textbf{Reference} \\
    \hline
    \multirow{2}{*}{Rule-based Model}     & Spelling Error Detection  & \multirow{2}{*}{\cite{PW-1,PW-2,PW-3,PW-4,PW-5,PW-6,PW-7,PW-8,PW-9,pybo,botok,thl-ewts,pw-22}} \\
    & Component Recognition & \\
    \hline
    \multirow{3}{*}{LSTM} & Spelling Error Detection  &   \cite{PW-10,PW-11} \\
   \cline{2-3} & Text Classification  &  \cite{PW-12}  \\
   \cline{2-3} & Component Recognition  &  \cite{PW-19}  \\
    \hline
    \multirow{2}{*}{CRT-based Model} & Tibetan Component Recognition &  \cite{PW-16,tispell}\\
    \cline{2-3}& Word Segmentation &  \cite{PW-13,PW-14,PW-15,PW-17,PW-20}  \\
    \hline
   Hybrid Model & Text Classification & \cite{PW-18} \\
    \hline
    \end{tabular}
\end{table}

As shown in Table~\ref{tai1}, we summarize the existing commonly used models. Since Tibetan language datasets are scarce and the language is a niche language, there are few related fields. The models used are basically the same. Especially BiLSTM-CRF\cite{PW-13,PW-14,PW-15,PW-17,PW-16,PW-20}, models in these papers are almost the same model with only minor adjustments to the parameters. Other models\cite{PW-1,PW-2,PW-3,PW-4,PW-5,PW-6,PW-7,PW-8,PW-9} are published in Chinese journal, but not open-source.

\noindent$\rightarrow$ \textbf{Document OCR \& Handwriting Recognition}: Tibetan OCR\cite{oct-1,ocr-2,ocr-3} and HWR systems\cite{tibetan-hwr-cnn,tibetan-hwr-dataset,tibetan-hwr-lexical} typically adopt a modular architecture combining page segmentation with neural sequence recognition. Character recognition is commonly implemented using CNN-RNN models with CTC decoding\cite{cnnrnn,PW-18,PW-15}, or Transformer-based attention recognizers\cite{attention,PW-18}, with lexicon-constrained decoding enforcing EWTS\cite{thl-ewts} and Unicode validity. Hybrid decoding workflows integrate language model priors to stabilize output under font distortion, handwritten variance, and stacked diacritic patterns. Practical pipelines include optional post-correction layers that apply grapheme normalization, syllable-level alignment, and lexical refinement to further reduce CER/WER and improve downstream usability\cite{tibetan-postcorr,tib-normalization,tib-lexical-refine}.

\noindent$\rightarrow$ \textbf{Machine Translation}: traditional Tibetan MT pipelines evolved from phrase-based statistical machine translation (SMT) to neural encoder-decoder architectures\cite{mt-1,mt-2,mt-3,mt-4,mt-5,mt-6,mt-7,mt-8}. Early systems adopted phrase extraction and alignment under Moses-style SMT, later replaced by RNN-based sequence-to-sequence models\cite{rnn,seq2seq} with attention. With the advent of the Transformer architecture, most Tibetan MT work shifted to Marian and fairseq-based Transformer baselines\cite{bert,attention}. Recent multilingual backbones such as mBART50\cite{bart}, mT5\cite{mt5}, and NLLB\cite{nllb} provide parameter-sharing across languages, enabling low-resource Tibetan translation through transfer learning and subword vocabulary alignment rather than fully supervised training. Although platforms like Google Translate\footnote{\url{https://translate.google.com/}} have recently introduced Tibetan support, their Tibetan translation performance remains shallow due to limited supervised data and the lack of Tibetan-specific linguistic priors, especially in religious, medical, and colloquial registers.

\noindent$\rightarrow$ \textbf{Speech Processing}: Tibetan ASR systems\cite{tibetan-asr-lmrescore,tibetan-asr-dialect,tibetan-tts-prosody,tibetan-tts-multispeaker} began with conventional hybrid HMM-DNN architectures\cite{kaldi}, later replaced by end-to-end CTC-based CNN-RNN\cite{cnnrnn} pipelines following DeepSpeech-style designs\cite{deepspeech}. With the emergence of self-supervised pretraining, wav2vec\,2.0\cite{wavenet} and Conformer-based models\cite{conformer} became the dominant architectures for Tibetan speech recognition, enabling low-resource adaptation via fine-tuning on limited Tibetan audio. For TTS, early parametric systems were replaced by Tacotron/Transformer-TTS\cite{tacotron} front-ends combined with neural vocoders such as WaveNet\cite{wavenet} or HiFi-GAN\cite{hifigan}, providing higher fidelity for multi-dialect Tibetan synthesis\cite{tmd,fmsd}.

\noindent$\rightarrow$ \textbf{Information Extraction \& Classification}: early Tibetan IE and classification tasks used lexical pattern matching and CRF-based sequence tagging\cite{crf}, followed by traditional machine learning classifiers such as SVM\cite{svm} and random forests\cite{rf} with handcrafted linguistic features\cite{svm-tibetan}. With the availability of multilingual encoders, BiLSTM-CRF architectures\cite{bilstmcrf} and fine-tuning of mBERT\cite{bert}, XLM-R\cite{xlmr}, or mT5\cite{mt5} became standard, allowing better generalization across domain-specific Tibetan terminology and named entities without requiring large manually engineered feature sets\cite{tibetan-ie-lexical,tibetan-crf-ie,tibetan-svm-cls,tibetan-bilstmcrf,tibetan-bert-ner,tibetan-domain-cls,tibetan-kg}
.

As model complexity increased across these individual tasks, research attention gradually shifted from task-specific architectures toward unified Tibetan pre-trained models, like LLMs\cite{r1,v3,gpt4o,llama,tibstccot,TIB-STC}, capable of performing OCR post-correction, translation, ASR transcription, and information extraction within a single instruction-following framework. 
This paradigm shift has led to the emergence of Tibetan-centric LLM initiatives that aim to consolidate multiple task abilities into a single generative backbone.
\begin{itemize}
    \item \textbf{Sun-Shine}\cite{TIB-STC}: the first Tibetan-centric LLM trained on the TIB-STC corpus via pretraining and supervised fine-tuning, targeting general Tibetan understanding and instruction following with unified tokenization and EWTS-compatible vocabulary design.
    \item \textbf{Sun-Shine 2.0}\cite{tibstccot}: an extended version incorporating CoT-style supervision, preference alignment, and safety filtering, enabling more robust reasoning, multi-turn dialogue, and controllable generation across Tibetan cultural, legal, and medical domains.
\end{itemize}
To the best of our knowledge, Sun-Shine\cite{TIB-STC} and its upgraded variant Sun-Shine 2.0\cite{tibstccot} currently represent the only two Tibetan-centric large language models explicitly designed with dedicated Tibetan pretraining, instruction tuning, and alignment objectives. For more details, please refer to Appendix \ref{app4}

For other pre-trained Tibetan language models\cite{peftt,TSBERT,ProSubCINO,tibert,cino-1,Tibetan-BERT-wwm}, existing efforts primarily focus on encoder-style architectures with Tibetan-adapted tokenizers\cite{llama2} or PEFT\cite{peft} strategies, which enhance representation quality for downstream tasks such as classification\cite{cino-1} and NER\cite{Tibetan-BERT-wwm} but do not offer the full generative and instruction-following capabilities of LLM-scale models.

\subsubsection{Memristor-Based Neuromorphic Tibetan AI}
To the best of our knowledge, no published work explicitly integrates memristor-based in-memory computing with Tibetan OCR, ASR, machine translation, or neural sequence modeling. Existing neuromorphic accelerators based on RRAM or PCM crossbar arrays have been primarily evaluated on Latin or Chinese character recognition\cite{mem-cnn-ocr,mem-hw-ocr}, generic sequence models for English NLP\cite{mem-nlp-accel,mem-seq2seq}, or language-agnostic Transformer inference\cite{mem-transformer-accel,mem-crossbar-attention}. These architectures demonstrate that convolutional filters, recurrent weight matrices, and attention projections can be mapped onto analog conductance states to achieve high-throughput MAC operations without digital memory shuttling. However, none of these systems account for the unique Unicode block structure, stacked diacritics, and syllable-centric token sparsity of Tibetan script, indicating a clear architectural gap between existing memristive accelerators and the linguistic requirements of Tibetan AI workloads.

Under this extension perspective, Tibetan-aware neuromorphic models can be derived by mapping script-specific convolutional kernels, CTC-style OCR decoders, and subword embedding matrices directly onto resistive crossbar arrays. For OCR, existing RRAM-based CNN accelerators\cite{mem-cnn-ocr,mem-hw-ocr} could be adapted to recognize Tibetan stacked glyph structures by encoding syllable-level visual primitives as conductance patterns. For text modeling, memristor LSTM and seq2seq accelerators\cite{mem-nlp-accel,mem-seq2seq} provide a direct analog to BiLSTM-style Tibetan segmentation models, while crossbar-based attention units\cite{mem-transformer-accel,mem-crossbar-attention} offer a hardware-aligned pathway for Tibetan subword attention computation originally implemented in mBART\cite{bert} or mT5-style\cite{mt5} architectures. This indicates that neuromorphic Tibetan AI is technically feasible through adaptation of existing analog MAC accelerators, but remains an unexplored research direction without task-specific architectural tailoring.

Recently, inspired by the feasibility of mapping Tibetan Unicode features to resistive states, Liu et al.\cite{liu2024tibetan,spellcheck} demonstrated a memristor crossbar array that performs syllable-component recognition by encoding Tibetan characters into binary and matching component patterns through in-memory current comparison. This work marks the first attempt to apply memristive computing to Tibetan script processing, indicating that hardware-centric approaches can be extended beyond generic OCR toward Tibetan-specific linguistic structures.

\clearpage

\section{Challenge}
\label{s5}

We address the challenges and discuss solutions to the adoption of AI in Tibetan.

\subsection{Data Scarcity}

In Tibetan AI, data scarcity encompasses not only the limited quantity of resources but also their restricted accessibility, narrow domain coverage, and lack of consistent annotation. Publicly available corpora are rare and largely confined to religious and literary texts, with minimal representation of modern domains such as news, science, or social media. Most resources are monolingual, while parallel corpora, speech data, and multimodal datasets are even scarcer. Rich linguistic annotations, such as tokenization, part-of-speech tags, syntactic dependencies, and named entities are limited, and inconsistencies in encoding standards and annotation schemes further hinder interoperability. These gaps include the absence of large-scale, domain-diverse monolingual corpora, high-quality parallel datasets, richly annotated resources for supervised AI tasks, multimodal alignments with speech or images, and standardized benchmarks for evaluation. The resulting shortage restricts model performance and generalization, limits the applicability of state-of-the-art architectures, narrows the range of downstream applications, and impedes reproducibility, thereby widening the technological gap between Tibetan and high-resource languages.
\\\textbf{$\to$ Potential Solution:} recent studies have explored multiple strategies to alleviate data scarcity. These include the application of data augmentation techniques such as back-translation, paraphrasing, and synthetic text generation; the use of transfer learning and cross-lingual pretraining from high-resource languages; community-driven data creation through crowdsourcing and academic collaborations; and domain adaptation approaches that incrementally fine-tune models on Tibetan-specific corpora. While each method offers partial mitigation, their combined adoption, together with the establishment of publicly accessible benchmarks, appears essential for sustainable progress in this low-resource language context.

\subsection{New Knowledge Adaptation}

A distinctive challenge in Tibetan AI is the adaptation to new knowledge arising from the language’s ongoing modernization. This process includes the growing digitization of Tibetan communication, the emergence of internet slang and neologisms, and lexical innovations influenced by contemporary Chinese, including semantic shifts and structural simplifications. Such developments generate a dynamic linguistic landscape in which vocabulary, orthography, and usage patterns evolve faster than traditional resources can capture. Without timely incorporation of these changes into corpora, lexicons, and models, AI systems risk becoming outdated, unable to accurately interpret contemporary expressions, or misaligned with real-world communication contexts.
\\\textbf{$\to$ Potential Solution:}
to address this challenge, continuous corpus updating and dynamic lexicon expansion are essential, leveraging web mining, social media monitoring, and community-driven annotation to capture emerging terms and usages. Domain-adaptive training, incremental fine-tuning, and terminology alignment with high-resource languages such as modern Chinese can facilitate rapid integration of new knowledge. Furthermore, establishing automated pipelines for detecting and validating neologisms would enable AI systems to maintain relevance in a rapidly evolving linguistic environment.

\subsection{Manual Calibration}

Manual calibration remains a persistent challenge in Tibetan AI due to the inherent difficulties of large-scale human verification. The process is constrained by the scarcity of qualified annotators, the considerable time and effort required to inspect vast quantities of data, and the variability introduced by inconsistent standards across different experts. Discrepancies in annotation guidelines, segmentation rules, and linguistic interpretations can lead to uneven quality and hinder reproducibility. The challenge is further compounded by the sheer volume of Tibetan text, particularly in projects involving large corpora or real-time data streams, making comprehensive manual review impractical.
\\\textbf{$\to$ Potential Solution:}
a promising approach is to develop standardized AI-based verification systems capable of performing the majority of calibration tasks, thereby reducing reliance on manual labor. Such systems can be trained on high-quality, expert-verified datasets to ensure alignment with agreed annotation standards, while incorporating active learning mechanisms to flag ambiguous or low-confidence cases for human review. In this hybrid workflow, the AI serves as a first-pass filter, enabling human annotators to focus on complex or contentious instances, thus improving both efficiency and consistency.

\subsection{Encoding Standard \& Inefficiency}

Encoding Tibetan script in digital environments presents unique challenges that affect both interoperability and processing efficiency. Although Unicode provides a standardized framework for Tibetan characters, inconsistencies in its adoption across software platforms, operating systems, and datasets often lead to compatibility issues. Historical reliance on legacy encodings and custom font systems has resulted in fragmented digital resources, where identical characters may be represented by different code points or rendered inconsistently. Such variability complicates text normalization, segmentation, and search, increasing preprocessing costs and reducing the reliability of downstream NLP tasks. These challenges are magnified at the tokenization and subword levels. Tibetan is written with syllable-based segmentation and complex stacked consonants, meaning that naive whitespace-based tokenizers often fail to identify meaningful boundaries. Subword-based methods such as Byte Pair Encoding (BPE) or SentencePiece can mitigate out-of-vocabulary issues but risk breaking apart semantically coherent syllables if not adapted to Tibetan orthography. Encoding inconsistencies, such as mixed use of legacy fonts and Unicode, can also cause identical syllables to be split into different tokens or merged incorrectly, thereby inflating vocabulary size, lowering model efficiency, and degrading performance.
\\\textbf{$\to$ Potential Solution:} a unified encoding pipeline that strictly adheres to Unicode normalization should be paired with Tibetan-specific tokenization and subword segmentation rules. This includes designing tokenizers that respect syllable boundaries while allowing controlled subword splits for rare terms. Open-source preprocessing tools should integrate automatic detection and conversion of non-standard encodings, alongside vocabulary optimization to reduce fragmentation. Collaboration between linguists, software engineers, and standards organizations is essential to maintain accurate, efficient, and interoperable Tibetan text processing across platforms.

\subsection{Ethical \& Safety Concern}

Development of AI systems for Tibetan raise complex ethical and safety considerations. From an ethical and moral standpoint, there is a need to ensure fairness, inclusivity, and respect for cultural and linguistic diversity, avoiding bias in datasets and preventing the marginalization of underrepresented dialects or communities. Politically and socially, language technologies may intersect with sensitive issues, including information security, censorship, and the potential misuse of AI for misinformation or surveillance, which can have implications for political stability and community trust. On the technical side, inadequate control over model outputs can lead to the generation of harmful, biased, or misleading content, while insufficient oversight of data sourcing can result in privacy violations or breaches of intellectual property.
\\\textbf{$\to$ Potential Solution:} addressing these concerns requires a multi-layered approach. This includes the establishment of transparent ethical guidelines for dataset creation and model deployment, adherence to privacy protection regulations, and the integration of bias detection and mitigation mechanisms into model training pipelines. Political and security risks should be assessed through impact reviews and controlled by access governance frameworks that regulate who can deploy and modify models. Continuous monitoring of model outputs, combined with culturally informed community oversight, can help maintain both technical safety and societal trust.

\clearpage

\begin{figure}[ht]
    \centering
    \includegraphics[width=\linewidth]{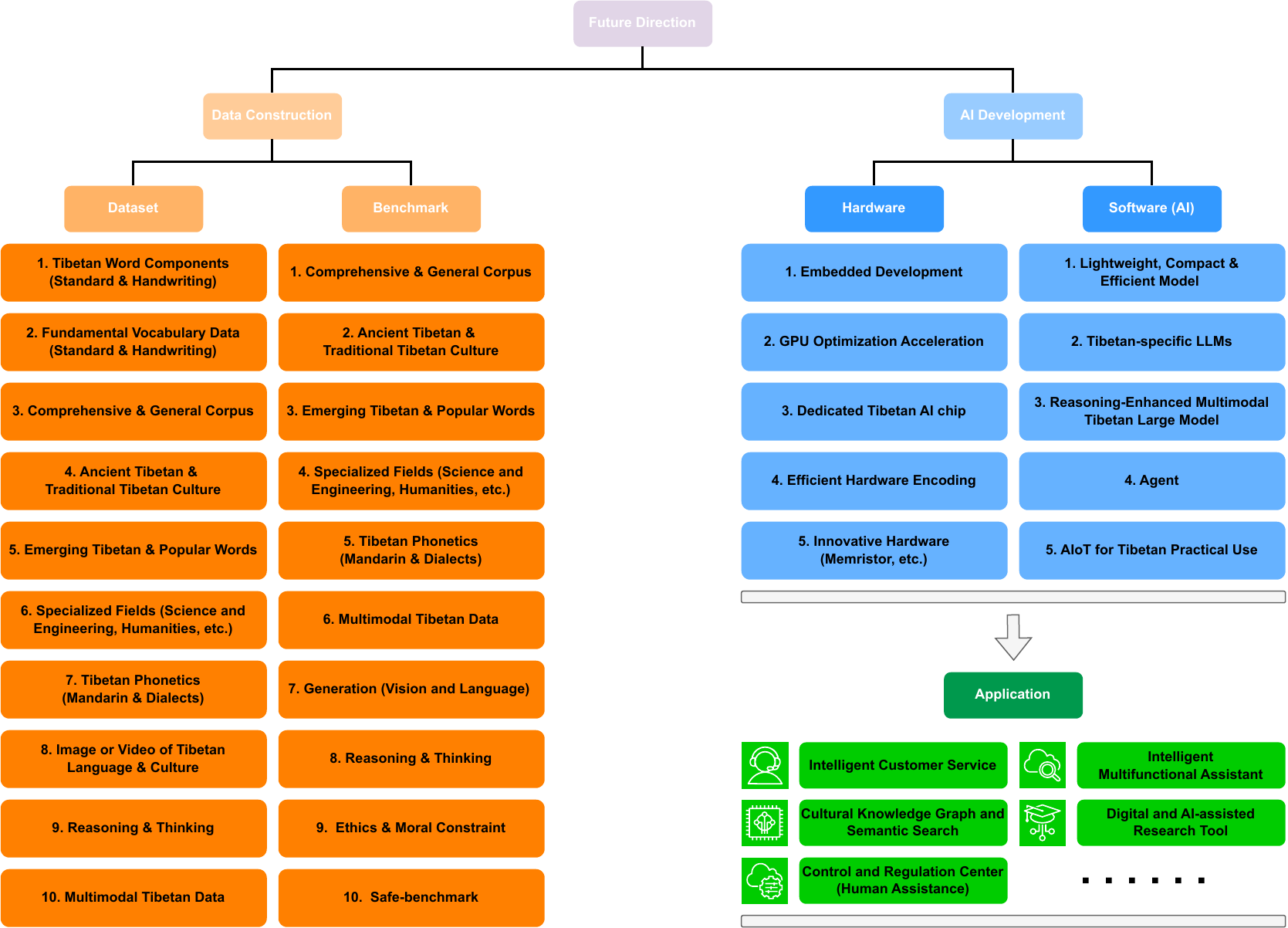}
    \caption{Future Directions of AI in Tibetan in terms of both Data and Models}
    \label{fd-1}
\end{figure}

\section{Future Direction}
\label{s6}

As shown in Figure~\ref{fd-1}, Tibetan AI is poised to evolve through coordinated progress in resource development, model and hardware innovation, and domain-specific applications, underpinned by strong ethical and governance frameworks. Building large-scale, high-quality, and standardized datasets and benchmarks remains the most pressing need, as they form the backbone for training and evaluating advanced systems. In parallel, innovations in algorithms and computational infrastructure will enhance efficiency and broaden the range of feasible applications. Finally, translating these technical achievements into socially beneficial and culturally sensitive solutions will be essential for sustainable impact.

\subsection{Data Construction}

\subsubsection{Dataset}

The construction of high-quality, task-specific datasets is the foundation of Tibetan AI development. At the most fundamental level, resources for basic tasks such as spelling correction, tokenization, and morphological analysis remain scarce, despite their essential role in enabling reliable downstream processing. Existing datasets are limited in size, often domain-specific, and rarely cover both standard printed Tibetan and handwritten script, which significantly constrains the robustness of AI models. Beyond these basic linguistic tasks, there is also a shortage of resources for higher-level natural language generation, including sentence and paragraph generation, summarization, and machine translation. Such datasets are critical for building general-purpose Tibetan language models, yet the available corpora are either too small or lack the diversity required to support robust training. At the highest level of complexity, the field faces a pronounced lack of multimodal datasets that align Tibetan text with speech, images, and videos capturing cultural, historical, and social contexts. These resources are indispensable for advanced tasks such as multimodal cultural heritage preservation, cross-modal retrieval, and reasoning-enhanced AI systems capable of integrating visual and linguistic information.

The scarcity of data is a bottleneck for both small-scale and large-scale models. Small models require carefully curated datasets to achieve acceptable accuracy under limited capacity, while large models rely on massive, diverse corpora to realize their full potential. Without systematic efforts to expand and standardize Tibetan datasets from basic spelling correction to complex multimodal cultural applications, the progress of Tibetan AI will remain fundamentally constrained.

\subsubsection{Benchmark}

Benchmark development provides the standardized foundation for evaluating Tibetan AI systems across linguistic, semantic, and multimodal tasks. While datasets enable model training, benchmarks ensure comparability, reproducibility, and progress tracking. In contrast to high-resource languages that benefit from mature suites such as GLUE\cite{glue}, XTREME\cite{xtreme}, MMLU\cite{mmlu} and CMMLU\cite{cmmlu}, Tibetan AI still lacks unified evaluation protocols and public benchmark sets.

Existing evaluations are fragmented, often limited to small, private test sets focused on specific domains, making it difficult to assess model generalization or cultural grounding. To address this, Tibetan benchmarks should evolve across multiple tiers:
\begin{itemize}
    \item Core Linguistic Benchmarks for Segmentation, Tagging, and Morphological Analysis
    \item General and Domain Benchmarks for Classification, Translation, and Semantic Understanding
    \item Cultural and Specialized Benchmarks covering Buddhist, Medical, and Legal Texts
    \item Multimodal Benchmarks for OCR, ASR, and Vision-language Generation
    \item Advanced Benchmarks for Reasoning, Ethics, and Safety Alignment
\end{itemize}

Developing such a hierarchical benchmark ecosystem will enable transparent model comparison, promote reproducible evaluation standards, and guide future progress toward culturally grounded, safe, and interpretable Tibetan AI systems.

\subsection{AI Development}

\subsubsection{Software}

The advancement of AI technologies tailored for Tibetan language processing will require innovations across model architectures, capabilities, and deployment strategies. At the foundational level, lightweight, compact, and efficient models are essential for ensuring accessibility in low-resource environments and enabling deployment on devices with limited computational capacity. Such models can facilitate the early adoption of AI in regions with restricted infrastructure while minimizing energy consumption. Building upon this foundation, the development of Tibetan-specific LLMs will be critical for capturing the unique linguistic, morphological, and cultural features of Tibetan. These models, trained on large-scale Tibetan corpora, will provide a robust basis for diverse downstream applications, from text understanding to dialogue systems. A further step will involve the creation of reasoning-enhanced multimodal Tibetan large models, integrating textual, visual, and auditory modalities with advanced inference capabilities. Such systems will enable complex cultural heritage preservation, cross-modal retrieval, and content generation that aligns with both linguistic structure and cultural context. Intelligent agents represent another promising direction, capable of autonomously performing tasks, interacting with users in Tibetan, and adapting to dynamic contexts. These agents could serve roles ranging from personalized learning assistants to domain-specific cultural guides.

Finally, AIoT systems designed for Tibetan practical use can connect AI capabilities with physical devices and real-world applications, supporting domains such as education, agriculture, tourism, and public administration. This integration will bring Tibetan AI from research laboratories into everyday life, ensuring that technological progress delivers tangible benefits to local communities.

\subsubsection{Hardware}

Advances in dedicated hardware will play a pivotal role in enabling efficient and scalable Tibetan AI systems. At the entry level, embedded development offers the possibility of deploying lightweight models on resource-constrained devices, making Tibetan language technologies accessible in low-connectivity or rural environments. Optimizing existing GPU architectures for Tibetan-specific workloads can significantly reduce training and inference latency, enabling faster experimentation and deployment without the need for extensive hardware investment. Beyond optimizing general-purpose hardware, future work may focus on the design of specialized Tibetan AI chips using electronic design automation tools. Such chips could incorporate dedicated instruction sets and processing units tailored for Tibetan script handling, including efficient support for complex character rendering and morphological processing. At the encoding level, hardware-oriented modifications, such as custom mappings that extend or adapt ASCII or Unicode representations for Tibetan, could further streamline data processing pipelines and reduce computational overhead.

At the frontier of hardware innovation, emerging devices such as memristors present new opportunities for high-density, low-power, and neuromorphic computing. Their ability to combine storage and computation within the same physical unit could be particularly advantageous for large-scale Tibetan AI applications that require both high-speed processing and energy efficiency. Progress along this trajectory from embedded systems and GPU optimization to custom AI chips, encoding-level innovation, and next-generation devices will define the hardware backbone of future Tibetan language processing technologies.

\subsection{Application}

The ultimate objective of Tibetan AI development lies in its translation into practical applications that deliver tangible benefits to society, culture, and research. Technological advances in data resources, AI models, and dedicated hardware will only reach their full potential when integrated into solutions that address real-world needs and contexts. Promising application areas include intelligent customer service systems capable of understanding and responding in Tibetan, thereby improving accessibility to information and services for Tibetan-speaking communities. Intelligent multifunctional assistants can serve as personalized guides, capable of performing a wide range of tasks such as language learning support, document translation, and daily information retrieval. Cultural knowledge graphs combined with semantic search technologies hold significant potential for preserving and promoting Tibetan heritage. These tools can enable scholars, educators, and the public to explore interconnected cultural, linguistic, and historical data through intuitive and intelligent interfaces. Digital and AI-assisted research tools can further enhance academic productivity by facilitating the analysis, annotation, and organization of Tibetan language and cultural materials.

In addition, control and regulation centers with human assistance can provide oversight for AI systems, ensuring safety, reliability, and ethical compliance in deployment. These applications represent concrete pathways for Tibetan AI to make a meaningful societal impact, highlighting that technological progress is ultimately measured by its ability to serve the needs of people, preserve cultural identity, and foster sustainable development.

\subsection*{Interdisciplinary Collaboration}

The advancement of Tibetan AI will increasingly depend on sustained interdisciplinary collaboration, integrating expertise from linguistics, computer science, cultural studies, education, and policy-making. Linguists provide critical insights into the structural and functional properties of Tibetan, including orthography, phonetics, and morphological variation across dialects, ensuring that AI systems are linguistically sound. Computer scientists contribute advanced algorithms, model architectures, and data processing pipelines that can address the technical challenges of low-resource language processing. Scholars in cultural and historical studies offer contextual knowledge that grounds AI applications in the lived realities, traditions, and heritage of Tibetan-speaking communities. Collaboration with the education sector is essential for developing AI-driven tools that support Tibetan language teaching, literacy programs, and curriculum design. Partnerships with policy-makers and legal experts can help establish regulatory and ethical frameworks that govern data collection, privacy protection, and cross-border data flows, ensuring compliance with both national and international standards. Engagement with local communities plays a central role, both in contributing authentic data and in shaping the priorities of AI projects to reflect real needs and values.

Interdisciplinary collaborations can be operationalized through joint research initiatives, shared resource platforms, and co-creation workshops that bring together diverse stakeholders. Such collaborations will not only accelerate technological innovation but also ensure that Tibetan AI solutions are culturally respectful, ethically grounded, and socially impactful.

\clearpage

\section*{Conclusion}

This survey provides the first systematic and large-scale review of Tibetan AI across linguistic resources, discriminative modeling, generative architectures, LLM adaptation, and emerging neuromorphic computation paradigms. Unlike high-resource languages with well-established benchmarks and data ecosystems, Tibetan AI remains fragmented, with isolated datasets, inconsistent annotation standards, limited cross-domain generalization, and a near-complete absence of hardware-efficient deployment strategies. Through structured categorization and comparative analysis, we identify the transition of Tibetan AI from rule-based and classical machine learning models to Transformer-based architectures and LLM-centric instruction-following systems, highlighting a paradigm shift toward unified generative modeling.

Despite this progress, key gaps remain: large-scale domain-diverse corpora are still scarce, reasoning and safety-aligned supervision for Tibetan is underdeveloped, and evaluation is constrained by the lack of standardized benchmarks beyond TLUE. Furthermore, current Tibetan LLMs are purely digital and GPU-dependent, overlooking the potential of neuromorphic accelerators such as memristor crossbar arrays, which are particularly suitable for low-resource settings and script-specific symbolic computation.

Looking forward, we argue that the development of Tibetan AI should evolve along three complementary axes: (1) \textbf{data-centric scaling} through culturally grounded corpus expansion and annotation standardization, (2) \textbf{alignment-centric modeling} through Tibetan-specific instruction tuning, safety filtering, and reasoning supervision, and (3) \textbf{hardware-centric deployment} via analog–digital co-designed architectures that leverage the discrete structure of Tibetan script for efficient token matching and inference. Advancing along these axes will not only improve the accessibility and inclusivity of Tibetan AI but also provide a transferable blueprint for other low-resource languages facing similar structural and computational constraints. We hope this survey serves as a foundation for building a sustainable and culturally aligned Tibetan AI ecosystem.

\section*{Acknowledgment}
This work was supported in part by the National Science and Technology Major Project under Grant 2022ZD0116100, in part by the Sichuan Provincial Major Science and Technology Project under Grant 2024ZDZX0012, in part by the National Natural Science Foundation of China under Grants 62276055, 62406062, 62436006 and 62406257,in part by the Sichuan Science and Technology Program under Grant 2023YFG0288, in part by the Natural Science Foundation of Sichuan Province under Grant 2024NSFSC1476, in part by the Tibetan Natural Science Foundation under GrantXZ202201ZR0054G.

\clearpage

\bibliography{bg}

\clearpage

\appendix

\section{Appendix: Proper Noun \& Key Parameter Abbreviation}
\label{app1}

As shown in Table~\ref{app-1}, some abbreviations also appear in other chapters; here, we list only those that occur for the first time within this chapter.

\begin{table}[h]
    \centering
        \caption{Proper Noun \& Key Parameter Abbreviation}
    \label{app-1}
    \scalebox{0.8}{
    \begin{tabular}{l|l|l}
    \hline
    \textbf{Part} & \textbf{Full Name} & \textbf{Abbreviation} \\
    \hline
   \multirow{3}{*}{Abstract} & Artificial Intelligence &  AI\\
    & Large Language Model & LLM  \\
    & Natural Language Processing  &  NLP\\
    \hline
    Section 1      &  -  & - \\
    \hline
    \multirow{9}{*}{Section 2}     &  Buddhist Digital Resource Center  & BDRC)  \\
          &  Tibetan and Himalayan Library   & THL \\
          &  Rangjung Yeshe Tibetan Corpus   & RYTC \\
          &  Tibetan Wikipedia Dump   & TWD \\
          &  Asian Classics Input Project  & ACIP \\
          &  OpenSubtitles Tibetan Fragment Collection  & OTFC \\
          &  Chain of Thought  & CoT \\
          & Tibetan Language Understanding Evaluation& TLUE \\
          & Tibetan Automatic Speech Recognition & ASR \\
    \hline
\multirow{18}{*}{Section 3} 
& Parameter-Efficient Fine-Tuning & PEFT \\
          & Word Segmentation & WS \\
          & Part-of-speech Tagging & POS \\
          & Named Entity Recognition & NER \\
          & Text Classification & TC \\
          & Sentence Similarity & SS\\
          & Entailment Detection & ED \\
          & Speech Recognition & SR \\
          & Optical Character Recognition & OCR \\
          & Machine Translation & MT \\
          & Statistical Machine Translation & SMT \\
          & Text Generation & TG \\
          & Summarization & SUM \\
          & Text-to-speech & TTS\\
          & Speech Synthesis & TTS \\
          & Question Answering & QA \\
          & Dialogue Generation & DG \\
          & Multimodal Generation & MG \\
    \hline
   \multirow{12}{*}{Section 4} &  Rule-based  & RB \\
    &  Machine Learning  & ML \\
    &  Deep Learning  & DL \\
    & Convolutional Neural Network    &  CNN \\
    & Recurrent Neural Network    &  RNN\\
    & Long Short-Term Memory    &   LSTM\\
          &   Machine Learning & ML \\
          &  Deep Learning  & DL \\
          & Support Vector Machine  &  SVM \\
          &  Hidden Markov Model  & HMM \\
          &  Conditional Random Field  & CRF \\
          &  Maximum Entropy Model  & MaxEnt \\
    \hline
     Section 5     &  -  & - \\
    \hline
    Section 6     &  -  & - \\
    \hline
    Appendix A   &  -  & - \\
    \hline
    \multirow{2}{*}{Appendix B}             & Reinforcement Learning from Human Feedback & RLHF \\
          & Direct Preference Optimization & DPO \\

    \hline
    Appendix D   &  -  & - \\
    \hline
    Appendix E   &  -  & - \\
    \hline
    \end{tabular}}
\end{table}

\clearpage

\section{Appendix: Details of TIBSTC \& TIBSTC-CoT}
\label{app2}

\subsection{TIBSTC: Structured Tibetan Corpus for Instruction-Tuning}

TIBSTC\cite{TIB-STC} is constructed as a large-scale structured Tibetan corpus with the explicit goal of serving as a foundation for Tibetan-centric language modeling. Unlike previous Tibetan datasets that primarily focused on isolated tasks such as OCR or machine translation, TIBSTC follows a unified representation scheme that organizes data at multiple linguistic levels, including syllable segments, word tokens, sentence pairs, and multi-turn instruction samples. All data is stored with parallel metadata fields for \texttt{Tibetan}, \texttt{Simplified Chinese}, and \texttt{English}, ensuring alignment consistency across downstream translation, classification, reasoning, and generation tasks.

\subsubsection{Corpus Sources \& Domain Coverage} The dataset aggregates Tibetan texts from four main categories: 
\begin{itemize}
    \item Digitized Classical Literature \& Canonical Buddhist Text
    \item Conversational Tibetan from Social Media \& Daily Communication
    \item Tibetan Legal \& Administrative Document
    \item Modern Tibetan News \& Educational Content
\end{itemize}

As reported in the original dataset release, around 66\% of the corpus comes from literature and religious texts, 24\% from contemporary web sources, and 10\% from media and documentary transcripts, providing both formal and colloquial linguistic coverage.

\subsubsection{Preprocessing and Normalization} All raw texts underwent Unicode normalization to remove incompatible codepoints while preserving Tibetan diacritic structures. To prevent the loss of traditional Tibetan morphology, no modern rewriting or aggressive spelling correction was applied; instead, non-standard syllable forms were preserved and annotated. Each data entry is packed into structured JSON format with hierarchical fields for \texttt{syllable\_segmentation}, \texttt{token\_alignment}, and \texttt{parallel\_translations}, enabling explicit control over syllable-level processing and cross-lingual supervision.

\subsubsection{Instruction Formatting} Following the principles of instruction-tuning corpora such as Alpaca\cite{alpaca} and FLAN\cite{flan}, TIBSTC introduces instruction-style prompts specifically adapted for Tibetan. Each prompt-entry pair contains a human-readable Tibetan instruction, a controlled-format system tag, and a target response. Unlike generic multilingual instruction datasets, TIBSTC enforces EWTS-compatible tokenization\cite{thl-ewts} to ensure that Tibetan script decomposition remains stable under byte-level tokenizers. The instruction subset is further divided into 4 subsets:
\begin{itemize}
    \item \textbf{Alpaca-Ti}: a Tibetan adaptation of Alpaca-style supervised instruction pairs, generated through translation-aligned prompting and manually verified to ensure that Tibetan imperative moods and honorific expressions remain semantically faithful to the original intent.

    \item \textbf{Safety-Prompts-Ti}: a curated subset focused on Tibetan safety alignment, containing refusal, ethical compliance, and content moderation prompts tailored to culturally sensitive Tibetan contexts such as religion, medical advice, and regional governance.

    \item \textbf{CValues-Ti}: a value-grounded instruction set incorporating Tibetan cultural priors, including compassion ethics, monastic discourse norms, and traditional knowledge frameworks, designed to steer model outputs toward culturally coherent Tibetan reasoning.

    \item \textbf{hh-rlhf-Ti}: a Tibetan variant of human-preference training pairs following HH-RLHF\cite{rlhf} conventions, where multiple Tibetan responses are ranked by native speakers to provide preference signals for RLHF\cite{rlhf} or DPO-style\cite{dpo} alignment in low-resource Tibetan settings.
\end{itemize}

They allow systematic scaling from supervised fine-tuning to preference alignment and safety modeling.

\subsubsection{Human Verification \& Linguistic Consistency} To ensure linguistic validity beyond automatic normalization, the dataset construction involved multiple rounds of manual auditing by native Tibetan speakers, including monks, legal clerks, and Tibetan computational linguists. Inconsistent segmentation or mistranslated parallel entries were annotated and filtered through majority judgment, providing higher semantic stability than web-scraped translation corpora.

\subsubsection{Dataset Role} Due to its structured alignment between script-level segmentation, multilingual mapping, and instruction semantics, TIBSTC functions not only as a dataset but also as a benchmark protocol. It is used as the base corpus for Tibetan-centric LLMs such as Sun-Shine\cite{TIB-STC} and later extended into the TIBSTC-CoT\cite{tibstccot} reasoning subset for multi-step Tibetan instruction alignment.

\subsection{TIBSTC-CoT: Chain-of-Thought Extension for Tibetan Reasoning Alignment}

While TIBSTC\cite{TIB-STC} provides token-level alignment and supervised instruction pairs for base model tuning, it does not explicitly encode multi-step reasoning or stepwise logical articulation. To address this limitation, TIBSTC-CoT\cite{tibstccot} is constructed as a reasoning-augmented extension that introduces explicit chain-of-thought supervision, enabling models to generate intermediate reasoning traces instead of producing only final answers.

\subsubsection{Multi-Agent CoT Generation Framework} TIBSTC-CoT follows a three-agent pipeline: (1) a question agent generates instruction prompts, (2) a reasoning agent produces multi-step explanation sequences, and (3) an evaluation agent assigns an automatic quality score on a 1-5 scale. 

Only samples with a score greater than or equal to 3.5 are preserved, after which human annotators perform manual quality refinement to ensure linguistic coherence and segmentation consistency.

\subsubsection{Reasoning Categories and Template Diversity} The CoT subset covers multiple reasoning types aligned with the TLUE benchmark\cite{tlue}, including comparison reasoning, legal interpretation, cultural logic grounding, analogy explanation, and multi-step factual inference. Each sample is formatted using an explicit structure with labeled reasoning steps (e.g., \textit{Step 1}, \textit{Step 2}, \textit{Therefore}), making the chain-of-thought trace directly learnable by generative Tibetan LLMs.

\subsubsection{Abstract Structure Illustration} Since this appendix focuses on structural representation rather than content disclosure, illustrating the abstract CoT template used throughout the dataset without including Tibetan text tokens.

\textbf{Abstract structure illustration:}
\begin{itemize}
    \item \textit{Instruction:} Explain the difference between two similar statements.
    \item \textit{Reasoning Trace:}
    \begin{itemize}
        \item Step 1: Identify semantic focus.
        \item Step 2: Compare logical implications.
        \item Step 3: Produce final explanation.
    \end{itemize}
    \item \textit{Final Answer:} Concise contrast statement.
    \item \textit{Score:} 4.3
\end{itemize}

\subsubsection{Integration with Model Alignment} TIBSTC-CoT is used as the primary chain-of-thought supervision source for Sun-Shine 2.0\cite{tibstccot}. The dataset is mixed with supervised instruction pairs following a 3:1 ratio, enabling models to learn both safe response behaviors and explicit reasoning traces. In subsequent evaluation, TIBSTC-CoT samples were shown to enhance reasoning-related performance categories in TLUE\cite{tlue}, improving consistency and interpretability of generated explanations compared to models trained without CoT alignment.

\clearpage

\section{Appendix: Details of TLUE Benchmark}
\label{app3}

\subsection{Overview}

\begin{figure*}[!ht]
  \centering
  \includegraphics[width=0.9\linewidth]{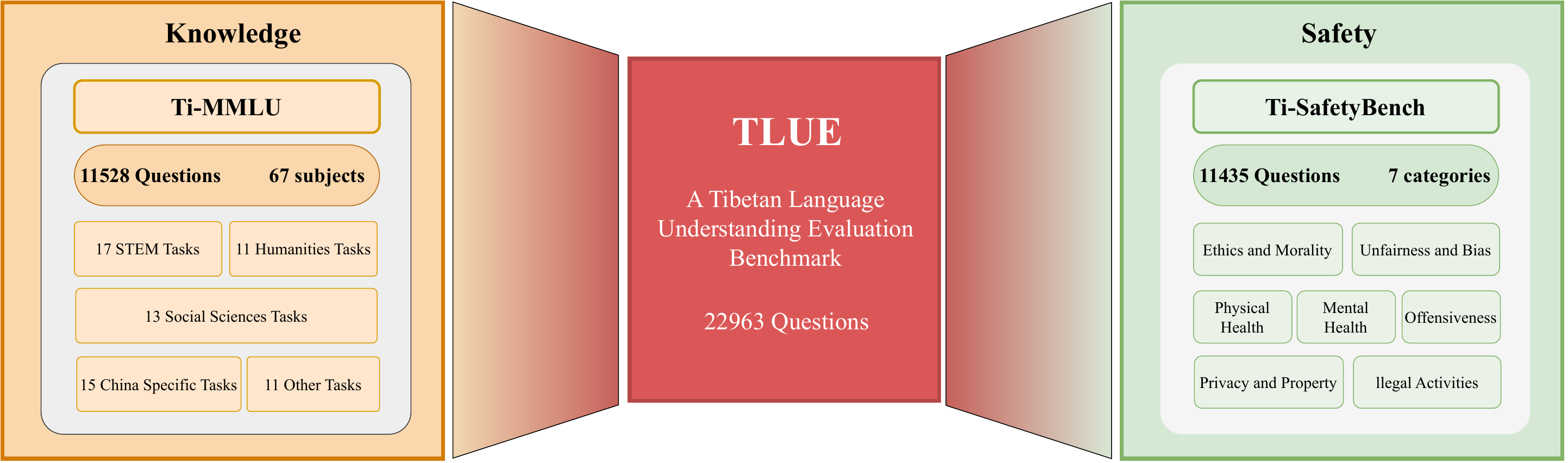} 
  \caption {Overview of the TLUE Benchmark}
  \label{fig:tlue_overview}
\end{figure*}

As shown in Figure~\ref{fig:tlue_overview}, TLUE provides a comprehensive evaluation suite for Tibetan language understanding by integrating both knowledge-oriented and safety-oriented assessment tracks. On the knowledge side, Ti-MMLU contains 11,528 questions across 67 subjects, covering STEM, humanities, social sciences, China-specific topics, and general knowledge, enabling systematic measurement of factual and reasoning capabilities in Tibetan. On the safety side, Ti-SafetyBench contributes 11,435 adversarial and alignment-sensitive questions spanning seven critical categories, including ethics, bias, health, offensiveness, privacy, and illegal content, targeting robustness and responsible model behavior. Together, these components form TLUE, a 22,963-question Tibetan Language Understanding Evaluation framework that jointly assesses both cognitive competency and value-aligned safety for Tibetan LLMs.

To the best of our knowledge, TLUE is the first and currently the only benchmark that systematically evaluates Tibetan language capabilities in LLMs. Proposed by our team, it has been accepted to the EMNLP 2025 main conference.

\subsection{Ti-MMLU}

For its sub-benchmark, Ti-MMLU\footnote{License: CC-BY-NC-SA 4.0}, it is an evaluation sub-benchmark designed specifically for Tibetan LLMs, similar to MMLU\cite{mmlu} in the English field and CMMLU\cite{cmmlu} in the Chinese field. It comprehensively tests the LLM's knowledge understanding and reasoning capabilities in a multi-disciplinary and multi-task environment through multiple-choice questions. 

\begin{figure*}[ht]
  \centering
  \includegraphics[width=1.0\linewidth]{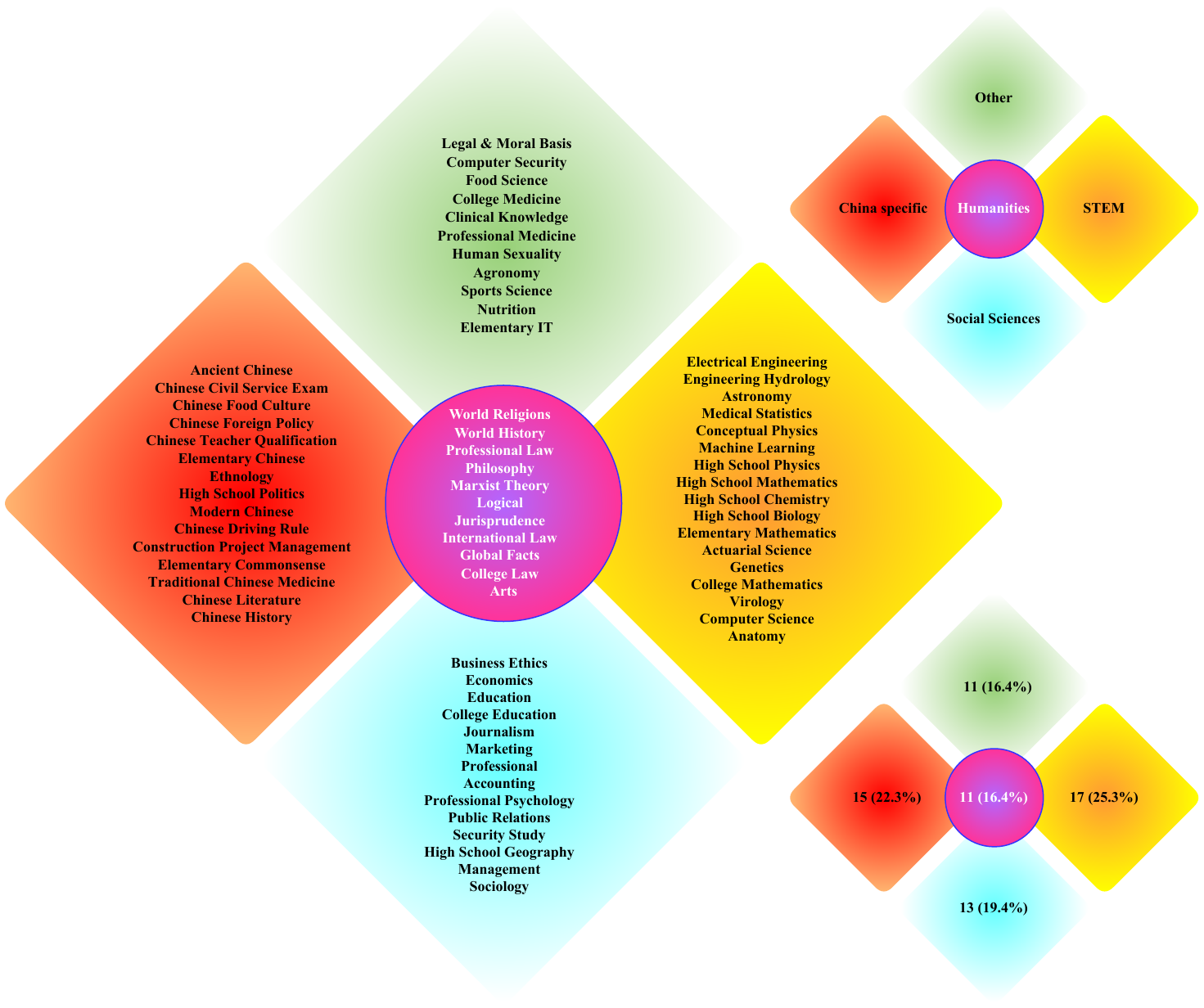} 
  \caption {Statistical Category of the Ti-MMLU Benchmark}
  \label{stat_ti_mmlu}
\end{figure*}

As shown in Firgure~\ref{stat_ti_mmlu}, Ti-MMLU contains 67 subtasks, covering multiple subject areas from middle school to university and even professional examinations, such as mathematics, physics, history, law, medicine, engineering, philosophy, literature, etc., covering the unique local knowledge system in Tibetan areas, such as college entrance examinations, teacher qualification certificates, medical examinations, etc., and is particularly suitable for evaluating the LLM's mastery of Tibetan language context and professional knowledge. 

Ti-MMLU uses zero-shot or few-shot settings, does not provide contextual learning, and directly examines the generality and true capabilities of the model. It is not only suitable for model comparison and ranking, but also helps developers discover the weak links of the model in specific fields, such as law and medicine.

\subsection{Ti-SafetyBench}

For its another sub-benchmark, Ti-SafetyBench\footnote{License: Apache License 2.0}, it is a multilingual benchmark systematically evaluate the safety of LLMs, similar to SafetyBench\cite{safetybench}, when handling sensitive or high-risk Tibetan content. It consists of multiple-choice questions, and supports zero-shot and few-shot evaluation settings to enable standardized comparisons across different models. 

Also, Ti-SafetyBench covers eight core safety categories: Offensiveness, Unfairness and Bias, Physical Health, Mental Health, Illegal Activities, Ethics and Morality, Privacy and Property, and Safety-Related Reasoning. These categories reflect common areas of concern in real-world applications where unsafe or harmful responses from LLMs may occur. 

By providing a structured, quantifiable framework for assessment, Ti-SafetyBench serves as a critical tool for developers, researchers, and policymakers in improving and regulating the deployment of safe and trustworthy AI systems for Tibetan.

\subsection{Summary}

Release of the TLUE benchmark addresses a critical gap in Tibetan AI: the lack of standardized evaluation settings. By offering the first domain-aware, task-diverse benchmark, TLUE lays the foundation for future research in Tibetan language understanding, especially in the context of multilingual and low-resource language modeling.

\clearpage

\section{Appendix: Sun-Shine LLM Family}
\label{app4}

\subsection{Sun-Shine: Base Tibetan Instruction Model}

Sun-Shine\cite{TIB-STC} is the first Tibetan-centric instruction-tuned language model trained exclusively on the TIBSTC\cite{TIB-STC} corpus. Rather than relying on multilingual pretraining, it adopts a tokenizer aligned with the EWTS writing convention to stabilize segmentation over Tibetan syllable structures. All instruction samples follow a controlled Alpaca-style formatting process with manual verification to ensure that directive semantics are preserved.

\subsubsection{Training Objective}
Supervised instruction tuning without any chain-of-thought decomposition or reinforcement-style preference optimization, focusing purely on direct input-to-output mapping.

\subsubsection{Corpus}
Trained exclusively on the TIBSTC instruction subset, which contains declarative prompt–response pairs but does not include multi-step reasoning traces or ranked candidate responses.

\subsubsection{Tokenizer}
Employs a custom EWTS-aligned vocabulary to ensure that Tibetan syllable units remain intact during subword segmentation, preventing fragmentation issues common in multilingual tokenizers.

\subsubsection{Interaction Format}
Restricted to single-turn instruction–response interactions without conversational memory or iterative refinement, reflecting a pure supervised fine-tuning regime.

\subsubsection{Alignment Scope}
No safety filtering, refusal modeling, or culturally grounded preference alignment signals are applied, making Sun-Shine a direct baseline of instruction-following behavior without alignment-level constraints.

\textbf{Alignment scope:} no safety filtering, refusal modeling, or culturally grounded preference alignment signals are applied, making Sun-Shine a baseline that reflects raw supervised instruction-following capabilities.

\subsection{Sun-Shine 2.0: CoT and Preference-Aligned Tibetan LLM}

Sun-Shine 2.0\cite{tibstccot} extends Sun-Shine by incorporating CoT supervision and preference-ranked alignment signals derived from TIBSTC-CoT, Safety-Prompts-Ti, and hh-rlhf-Ti. In contrast to direct answer generation, the model is trained to produce multi-step reasoning traces using labeled inference stages such as \textit{Step 1}, \textit{Step 2}, and \textit{Therefore}.

\subsubsection{Training Objective}
Combined CoT enhanced supervised fine-tuning and preference-based alignment following a DPO-style optimization objective, enabling the model to internalize both direct instruction mapping and ranked response preference patterns.

\subsubsection{Reasoning Supervision}
Exposed to multi-step reasoning traces derived from TIBSTC-CoT, where explicit stepwise inference structures are embedded to encourage interpretable intermediate output generation instead of single-pass answers.

\subsubsection{Alignment Mechanism}
Trained with preference-ranked response pairs aligned to RLHF\cite{rlhf}/DPO\cite{dpo} conventions, allowing the model to distinguish between acceptable and suboptimal Tibetan responses based on human- or model-evaluated preference signals.

\subsubsection{Safety Scope}
Injected with structured refusal and compliance prompts to regulate ungrounded generation, establishing controllable output behavior under safety, neutrality, and culturally coherent response constraints.

\subsubsection{Interaction Format}
Supports stepwise reasoning expansion and more stable multi-turn dialogue behavior, in contrast to the strictly single-turn instruction–response interface of the initial Sun-Shine model.

\clearpage

\section{Appendix: Memristor \& Tibetan AI}
\label{app5}

\subsection{Motivation: Beyond Digital Tibetan NLP}
Conventional Tibetan NLP systems, including OCR, ASR, spellchecking, and LLM-based reasoning, rely entirely on digital GPU computation. While effective, these architectures suffer from repeated memory shuttling between compute units and storage, especially when processing syllable-structured tokens that require frequent pattern matching and subword alignment. Memristor-based in-memory computing offers a fundamentally different pathway by enabling matrix-vector multiplications and pattern similarity operations directly within resistive crossbar arrays. For languages such as Tibetan, where token discrimination depends on a limited set of combinatorial syllable components, this analog computation model aligns naturally with the structure of the script.

\subsection{Existing Hardware Prototypes for Tibetan Symbolic Processing}
Only two published hardware efforts currently exist in this direction. The first demonstrates a memristive computing architecture for Tibetan syllable component recognition, where binary-encoded component patterns are projected onto memristor arrays for in-memory matching\cite{liu2024tibetan}. The second work extends this idea toward spell-level error detection using a similar crossbar-based current comparison mechanism, marking the first use of analog resistive hardware for Tibetan orthographic processing\cite{spellcheck}. Both systems operate on symbolic representations rather than neural embeddings, but they provide concrete evidence that Tibetan-specific pattern structures can be mapped into analog memory states.

\subsection{Architectural Relevance to Future Neuromorphic Tibetan LLMs}
Although current memristor-based Tibetan accelerators focus on discrete rule-based matching, their structural properties mirror fundamental operations in LLM inference, such as embedding projection, token gating, and attention-based similarity scoring. The resistive crossbar mapping used for syllable detection can be viewed as a hardware-level analogue of subword embedding lookup or prefix gating during decoding. This suggests that future Tibetan LLMs could be partially offloaded to specialized analog arrays that accelerate token activation and linguistic pattern filtering before digital transformer layers perform high-level reasoning.

\subsection{Toward Hybrid Analog–Digital Tibetan Language Models}
These early prototypes indicate that Tibetan AI workloads possess a discrete symbolic structure that makes them suitable for hybrid analog–digital architectures. A feasible future research path is to integrate memristor-based prefix and subcomponent matching units as a pre-filter stage in LLM decoding, allowing analog hardware to discard low-relevance token candidates before digital attention computation. This would reduce memory bandwidth pressure and align with the low-resource setting of Tibetan NLP, where efficiency gains can directly improve accessibility in edge devices and offline educational deployments.

\end{document}